\documentclass[letterpaper]{article} 
\usepackage{aaai2026}  
\usepackage{times}  
\usepackage{helvet}  
\usepackage{courier}  
\usepackage[hyphens]{url}  
\usepackage{graphicx} 
\usepackage{natbib}  
\usepackage{caption} 
\usepackage{algorithm}
\usepackage{algorithmic}

\usepackage{selectp}

\usepackage{newfloat}
\usepackage{listings}
\DeclareCaptionStyle{ruled}{labelfont=normalfont,labelsep=colon,strut=off} 
\floatstyle{ruled}
\newfloat{listing}{tb}{lst}{}
\floatname{listing}{Listing}
\usepackage{amsmath}
\usepackage{tcolorbox}
\usepackage{booktabs}
\usepackage{multirow}
\usepackage[capitalize]{cleveref}
\usepackage{xspace}
\usepackage{rotating}
\usepackage[shortlabels,inline]{enumitem}
\usepackage{amssymb}
\usepackage{soul}
\usepackage{tabularx}
\usepackage[font=footnotesize,skip=3pt,subrefformat=parens]{subcaption}
\usepackage{mathtools}

\usepackage{xcolor}

\newcommand{\mytiny}[1]{\text{\fontsize{8.7pt}{9.2pt}\selectfont #1}}

\makeatletter
\DeclareRobustCommand\onedot{\futurelet\@let@token\@onedot}
\def\@onedot{\ifx\@let@token.\else.\null\fi\xspace}
\def\eg{\emph{e.g}\onedot} 
\def\ie{\emph{i.e}\onedot}

\makeatother

\title{ImageSet2Text: Describing Sets of Images through Text}
\author{
    Piera Riccio\equalcontrib\textsuperscript{\rm 1}\footnotemark[2], Francesco Galati\equalcontrib\textsuperscript{\rm 2}, Kajetan Schweighofer\textsuperscript{\rm 3}, Noa Garcia\textsuperscript{\rm 4}, Nuria Oliver\textsuperscript{\rm 5}
}
\affiliations{
    \textsuperscript{\rm 1}University of Amsterdam \textsuperscript{\rm 2}Independent Researcher \textsuperscript{\rm 3}Johannes Kepler University Linz\\
    \textsuperscript{\rm 4}The University of Osaka \textsuperscript{\rm 5}ELLIS Alicante\\
    p.riccio@uva.nl
}

\definecolor{desertlavender}{HTML}{D9D2E9}

\begin{document}

\twocolumn[{%
\renewcommand\twocolumn[1][]{#1}%
\maketitle

\begin{center}
    \includegraphics[width=0.95\textwidth]{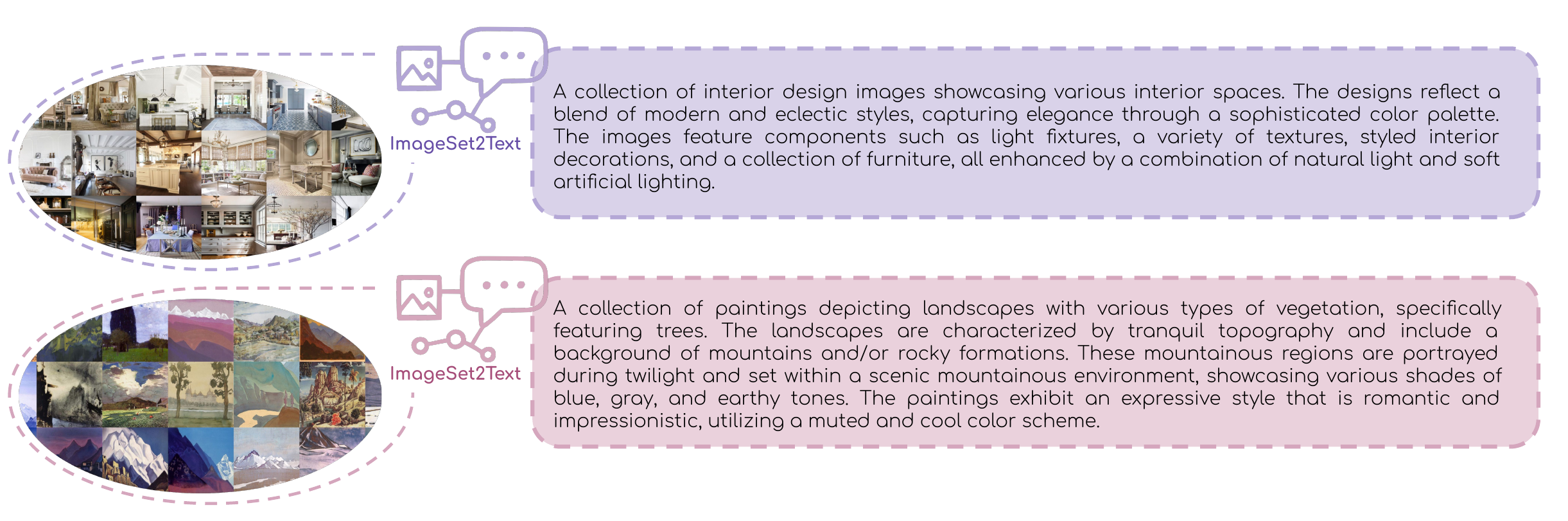}
\captionof{figure}{
\texttt{ImageSet2Text} generates detailed and nuanced descriptions from large sets of images. We report exemplary generated descriptions for two groups of images \cite{sharma2018, artgan2018}.}
\label{fig:fig1}
\end{center}%

}]

\renewcommand{\thefootnote}{\fnsymbol{footnote}}
\footnotetext[1]{Equal contribution.\quad\textsuperscript{\dag}Work conducted at ELLIS Alicante.}
\renewcommand{\thefootnote}{\arabic{footnote}}

\begin{abstract}
In the era of large-scale visual data, understanding collections of images is a challenging yet important task. To this end, we introduce \texttt{ImageSet2Text}, a novel method to automatically generate natural language descriptions of image sets.  
Based on large language models, visual-question answering chains, an external lexical graph, and CLIP-based verification, \texttt{ImageSet2Text} iteratively extracts key concepts from image subsets and organizes them into a structured concept graph. 
We conduct extensive experiments evaluating the quality of the generated descriptions in terms of accuracy, completeness, and user satisfaction. We also examine the method's behavior through ablation studies, scalability assessments, and failure analyses. Results demonstrate that \texttt{ImageSet2Text} combines data-driven AI and symbolic representations to reliably summarize large image collections for a wide range of applications. 
\end{abstract}

\begin{links}
     \link{Code}{https://github.com/ellisalicante/ImageSet2Text}
     \link{Datasets}{code + /tree/main/data}
    \link{Extended version}{https://arxiv.org/pdf/2503.19361}
 \end{links}

\section{Introduction}
\label{sec:intro}

The analysis of large-scale image sets is essential to uncover visual patterns that isolated samples fail to reveal \cite{deng2025visual}. For example, describing a dataset of historical newspaper photographs can disclose stylistic trends over time or systematic biases (such as underrepresented demographics or recurring stereotypical depictions), which only become apparent when analyzing the images collectively. To make such analyses scalable to large image collections, automated tools are indispensable. While visualizations help identify broad patterns \cite{manovich2012}, automatically generated textual summaries can transform overwhelming visual collections into \textit{interpretable} knowledge.

A variety of application domains would benefit from automatic summarization of image collections, including assistive technologies \cite{bigham2010, gurari2020}, cultural analytics \cite{shen2019discovering,cetinic21, manovich2020}, bias detection \cite{khosla2012undoing}, socio-economics \cite{jean2016combining,dubey2016deep,deng2025visual}, exploratory data analysis \cite{boiman2007detecting}, and training data transparency \cite{gebru2021} --- whose need is intensified by emerging AI regulations \cite{EU_AI_Act_2024}. In addition, in explainable AI, dataset-level insights have been found to be valuable for influential sample analysis and data segmentation \cite{park2023, shah2023, chung2019, d2022, eyuboglu2022}. 

However, while large vision-language models have made significant progress in the last few years \cite{achiam2023,bai:2025}, their application to describing large-scale image collections remains limited. Current approaches typically handle either individual images for captioning \cite{hossain2019, vinyals2015, xu2015} or descriptions of small curated sets \cite{Chen_2018, alayrac2022, li2023mimic, yao2022}, failing to address the challenges of summarizing large visual datasets. This limitation remains unresolved due to fundamental technical challenges in processing multiple visual inputs at once \cite{visdiff,deng2025visual}. 

In this paper, we propose \texttt{ImageSet2Text}, a method for generating natural language descriptions\footnote{We refer to captions as “short pieces of text” \cite{cambridge}, while descriptions are longer and more detailed.} \cite{pi2024} of large collection of images, as shown in Fig. \ref{fig:fig1}. \texttt{ImageSet2Text} leverages multimodal large language models (LLMs) through an iterative visual question answering (VQA) process, which combines hypothesis formulation-verification with external knowledge to extract comprehensive insights from image collections. Inspired by recent concept bottleneck models (CBMs) \cite{koh2020,tan2024,chattopadhyay2024}, \texttt{ImageSet2Text} extends this interpretable approach based on intermediate open-set concept prediction beyond classification tasks. 

The descriptions generated by \texttt{ImageSet2Text} are obtained by means of an iterative process, consisting of two phases: (a) \textit{Guess what is in the set}, and (b) \textit{Look and keep}. To ensure scalability, a small subset of images is randomly selected in each iteration. In \textit{Guess what is in the set}, an LLM-based VQA process identifies prevalent visual elements within the subset, and an external lexical graph is integrated to formulate hypotheses about the original set of images. In \textit{Look and keep}, the hypotheses are verified using contrastive vision-language (CVL) embeddings \cite{radford2021} to assess their consistency across the entire image set. Verified hypotheses are joined into a \textit{concept graph} and used to seed the next iteration. The process terminates when the concept graph cannot be updated anymore, at which point a final description of the entire image set is generated based on the accumulated information.

\begin{figure}[t]
    \centering
    \includegraphics[width=\linewidth]{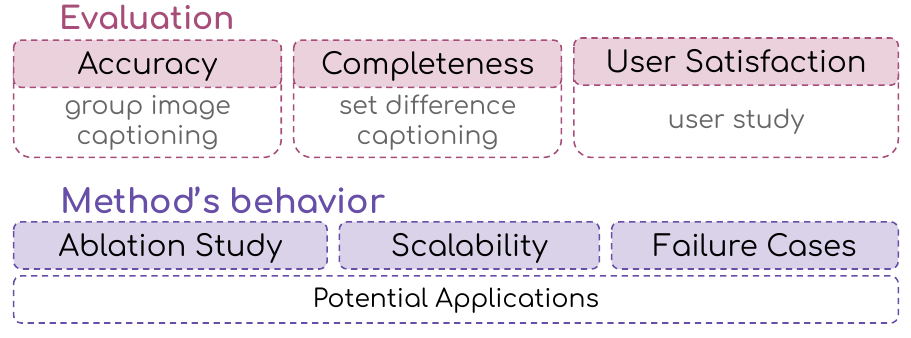}
    \caption{\texttt{ImageSet2Text}'s evaluation considers three key properties of the descriptions (accuracy, completeness and user satisfaction) and analyzes the method's behavior through ablations, scalability estimates and failure cases analysis, showing great versatility for potential applications.}
    \label{fig:evaluation_pipeline}
\end{figure}

In extensive experiments (see Fig. \ref{fig:evaluation_pipeline}), we evaluate \texttt{ImageSet2Text}´s descriptions 
according to their: 1) \emph{accuracy}, via a large-scale group image captioning experiment; 2) \emph{completeness}, by means of an image sets comparison task; and 3) \emph{user satisfaction}, through a user study. We also evaluate its behavior via: 4) an \textit{ablation study}, 5) a \textit{scalability estimate}, and 6) an \textit{analysis of failure cases}. Our results indicate that the generated descriptions accurately capture the visual content of large collections, offering rich detail and human-friendly readability, highlighting the potential of \texttt{ImageSet2Text} for diverse applications. 

\section{Related Work}
\label{sec:related}

\paragraph{Image Captioning} generates short textual descriptions of images by recognizing objects, attributes, scenes, and their relationships \cite{hossain2019}. Early methods relied on deep learning for feature extraction \cite{vinyals2015, xu2015}, while recent approaches leverage LLMs. For example, ChatCaptioner integrates VQA with chat logs for iteratively refinement \cite{zhu2023}, and \citeauthor{mao2024}~(\citeyear{mao2024}) proposes the creation of context-aware, user-specific captions. The importance of context has been highlighted in image captioning for Art History, where different interpretations can lead to different descriptions \cite{bai21,cetinic21,yue24}.

\paragraph{Group-Image Captioning} extends single-image captioning to small sets (typically $2-30$ images) by identifying shared patterns in the images \cite{Chen_2018, alayrac2022, li2023mimic, yao2022}. Proposed methods include modeling temporal relationships among images \cite{wang2019}, analyzing pairwise differences  \cite{chang2023, kim2021, park2019}, and comparing target and reference groups of images \cite{li2020}. Scene graphs have also been used to model and summarize the relationships between visual elements \cite{phueaksri2024,phueaksri2023}, while LLMs appear to be promising for both individual and small-group captioning \cite{achiam2023}. Benchmarks focus on spatial, semantic, and temporal aspects of small groups of images \cite{meng2024} or on evaluating large vision-language models in multi-image question answering \cite{liu2024}. Despite the variety of approaches, the main limitation remains scaling to larger groups containing hundreds or thousands of images \cite{phueaksri2023}.

\paragraph{Understanding Collections of Images} remains challenging despite its importance in today's world of large-scale visual data. Existing approaches, such as textual PCA \cite{hupert2022}, concept-level prototypes \cite{doersch2015, van2023}, color-based statistical analysis \cite{torralba2011}, and set-level classification \cite{wang2022} fail to produce interpretable textual descriptions. Steps towards bridging this gap 
include set difference captioning (SDC) \cite{visdiff} and temporal change detection in urbanist images \cite{deng2025visual}. In this paper, we advance beyond comparative approaches by proposing a novel method for generating comprehensive textual descriptions of large image collections. 

\paragraph{Foundation Models} are increasingly used for complex vision-language tasks. In addition to SDC \cite{visdiff}, querying a VQA model through an LLM has been used to iteratively improve image and video captions \cite{chen2023, zhu2023}, to detect biases in text-to-image generation \cite{d2024}, and to evaluate text-to-image generation faithfulness \cite{hu2023}. \texttt{ImageSet2Text} extends this paradigm by integrating multiple foundation models to describe image collections, contributing to emerging research \cite{deng2025visual}.

\section{\texttt{ImageSet2Text}}
\label{sec:method}

\begin{figure*}[!ht]
    \centering
    \includegraphics[width=\linewidth]{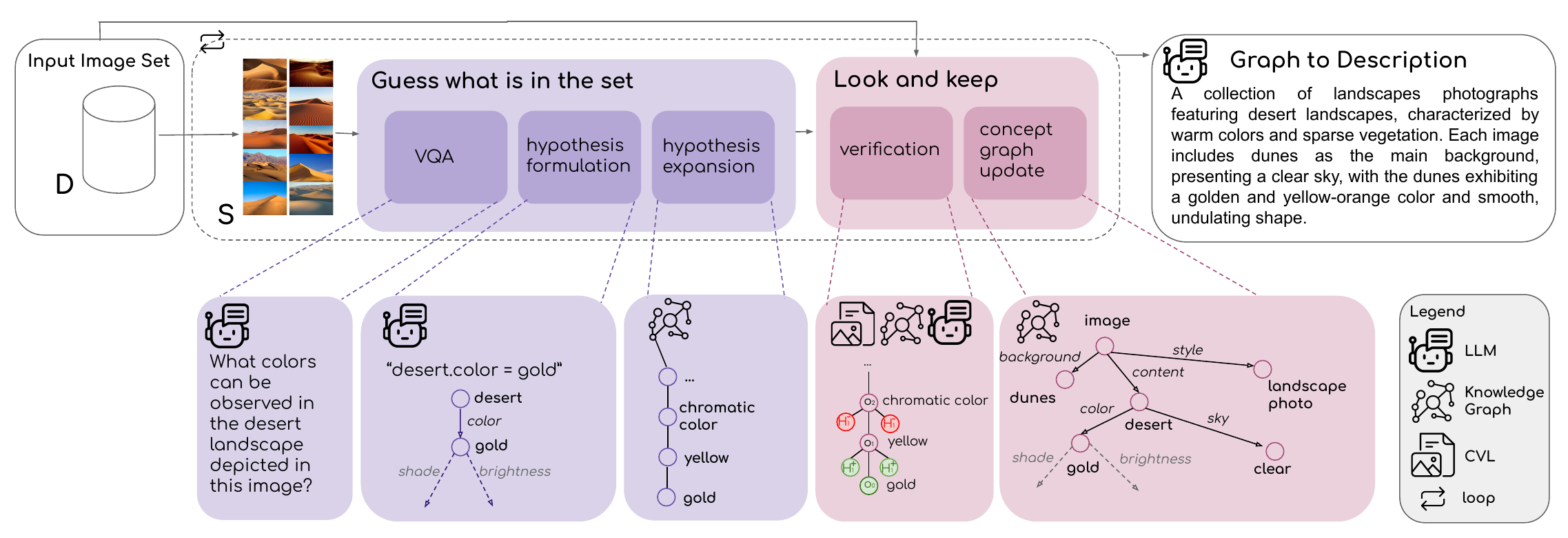}
    \caption{Overview of \texttt{ImageSet2Text}, considering an example set from the PairedImageSets datasets \cite{visdiff}. The figure shows how the different modules of the iterative process allow inferring information from the input image set, eventually generating a nuanced textual description.}
    \label{fig:sketch}
\end{figure*}

\texttt{ImageSet2Text} generates textual descriptions of image sets that highlight the common visual elements present in most of the images. As shown in \cref{fig:sketch}, it leverages external prebuilt components, including an LLM, a VQA model, a lexical graph, and a CVL model, to construct an intermediate concept graph, which then serves as the basis for generating the description. Given a set of \( N \) images {\( \mathcal{D} = \{x_1, \dots, x_N\} \)}, \texttt{ImageSet2Text} automatically generates a textual description $d$ that summarizes the visual elements in $\mathcal{D}$. This is achieved by constructing an intermediate concept graph represented as a list of verified hypotheses $\mathcal{G}_c = \{ h_\checkmark^1, ..., h_\checkmark^T \}$, where each $h_\checkmark^\tau$ is a triplet $\langle s, p, o \rangle$ identifying a subject $s$, a predicate $p$, and an object $o$ that capture the key visual elements and their relations in $\mathcal{D}$. To build $\mathcal{G}_c$, \texttt{ImageSet2Text} follows an iterative process with $T$ iterations depicted in \cref{fig:sketch}, with the following steps: 
\begin{enumerate}
\item \textbf{Initialization} ($\tau = 0$): It starts by defining the first subject $s \coloneq \text{`\textit{image}'}$ and a list of three candidate predicates, $P \coloneq \{\text{`\textit{content}'},\;\text{`\textit{background}'},\;\text{`\textit{style}'}\}$. The initial $\mathcal{G}_c^0$ contains only `\textit{image}' as root node.

\item \textbf{Iterations} ($\tau = 1, ..., T-1$), composed of two phases:

    \begin{enumerate}
        \item \textbf{Guess what is in the set} – A random  subset of images \( \mathcal{S} \subset \mathcal{D} \), with \( |\mathcal{S}| = M \ll N \), is analyzed to hypothesize a full triplet $\langle s, p, o \rangle$ of elements present in $\mathcal{D}$, where $M$ is a predefined parameter.
        
        \item \textbf{Look and keep} – The formulated hypothesis is verified on \( \mathcal{D} \). If confirmed, it is used to update $ \mathcal{G}_c^\tau$.
        
    \end{enumerate}

\item \textbf{Termination} ($\tau = T$): 
After convergence at $\tau= T$, a coherent textual description \( d \) is generated from the final graph representation \( \mathcal{G}_c = \mathcal{G}_c^T \).
\end{enumerate}

\noindent Next, we describe the two phases of the iterations.

\subsection{Guess what is in the Set}
The first phase generates a hypothesis from the random sample $\mathcal{S}$ of $M$ images for later verification on the full set $\mathcal{D}$. Let $\tau$ denote the current time step. From the current graph $\mathcal{G}_c^\tau$, \texttt{ImageSet2Text} selects the first encountered closest leaf node to the root node as the subject $s$, along with a random candidate predicate $p \in P$ appropriate for $s$. \cref{fig:sketch} depicts an illustrative example where the image set \(\mathcal{D}\) contains desert photos. At iteration $\tau=5$, the subject selected is $s = \text{`\textit{desert}'}$ and the predicate assigned is $p = \text{`\textit{color}'}$.

\paragraph{VQA.} An LLM is prompted to ask a specific question about the images in $\mathcal{S}$ depending on the values of $s$ and $p$. In the given example, with $s = \text{`\textit{desert}'}$ and $p = \text{`\textit{color}'}$, the resulting question is:

\begin{tcolorbox}[colback=desertlavender, colframe=black!70, boxrule=1pt, arc=1mm]
\begin{center}
\textit{What colors can be observed in the desert landscape depicted in this image?}
\end{center}
\end{tcolorbox}

The question is posed to a VQA model for each of the $M$ images \( x_i \in \mathcal{S} \), resulting in a set of answers \( \mathcal{A} = \{a_1, a_2, \dots, a_M\} \), where  \( a_i \) denotes the answer for $x_i$. 

\paragraph{Hypothesis formulation.} 
Given \(\mathcal{A}\), each iteration verifies whether the predicate $p$ can expand $\mathcal{G}_c^\tau$. To achieve this, the LLM is prompted to perform two tasks: \textbf{1)~summarization}, \ie, condense $\mathcal{A}$ into a single hypothesis $h_0$, which is formulated as a triplet where the subject $s$ and predicate $p$ are given, and the object $o_0$ has to be derived from \(\mathcal{A}\); and \textbf{2)~completion}, \ie, suggest possible expansions of $\mathcal{G}_c^\tau$ from $o_0$ as a list of new predicates to append to $P$. Following our example, \(\mathcal{A}\) would yield hypothesis $h_0$ and continuations $P$:

\begin{tcolorbox}[colback=desertlavender, colframe=black!70, boxrule=1pt, arc=1mm]
\begin{center}

\(h_0 = \langle \text{`\textit{desert}'}, \text{`\textit{color}'}, \text{`\textit{gold}'} \rangle \)

$P \leftarrow \{\text{`\textit{shade}'}, \text{`\textit{brightness}'}\}
$
\end{center}

\end{tcolorbox}

\paragraph{Hypothesis expansion.} Since $h_0$ is derived from the subset $\mathcal{S}$, it may not generalize well to the full image set $\mathcal{D}$ due to sampling bias. To mitigate this, \texttt{ImageSet2Text} creates a set \( \mathcal{H} = \{h_0, h_1, \dots, h_k\} \) of hypotheses ordered from the most general (\( h_k \)) to the most specific (\( h_0 \)), where $h_k \supset h_{k-1} \supset \dots \supset h_0$. To generate \( \mathcal{H} \) from $h_0$, \texttt{ImageSet2Text} relies on a given lexical graph $\mathcal{G}_l$. Let $\mathcal{G}_l = (V, R)$ be a directed graph, where \( V \) is the set of lexical entries and \( R \) represents semantic relations between nodes in \( V \). For any given node \( v \in V \) (\eg, `\textit{gold}'), its parent node denotes a more general concept, or hypernym (\eg, `\textit{yellow}'), and vice versa: `\textit{gold}' is a hyponym of `\textit{yellow}', \ie, a more specific lexical concept. In addition, two nodes \( v_1, v_2 \in V \) are sibling nodes if they share the same parent node, but correspond to different lexical concepts (\eg, `\textit{gold}' and `\textit{gamboge}', both hyponyms of `\textit{yellow}').

The set \( \mathcal{H} \) is obtained by traversing upward in the knowledge hierarchy of $\mathcal{G}_l$ by a maximum number of steps \( \delta \). Given a hypothesis \( h_i = \langle s, p, o_i \rangle \), its generalization \( h_{i+1} \) is created as \( h_{i+1} = \langle s, p, o_{i+1} \rangle \), where $o_{i}=\text{parent}^i(o_0)$ and the parent function is the operation of moving to the hypernym of a lexical entry in $\mathcal{G}_l$. In the ongoing example, the hypotheses in $\mathcal{H}$ follow the hierarchy: 

\begin{tcolorbox}[colback=desertlavender, colframe=black!70, boxrule=1pt, arc=1mm]

\begin{center}
\(h_0 = \langle \text{`\textit{desert}'}, \text{`\textit{color}'}, \text{`\textit{gold}'} \rangle \)

\(h_1 = \langle \text{`\textit{desert}'}, \text{`\textit{color}'}, \text{`\textit{yellow}'} \rangle \)

\(h_2 = \langle \text{`\textit{desert}'}, \text{`\textit{color}'}, \text{`\textit{chromatic color}'} \rangle \)
\end{center}

\end{tcolorbox}

\subsection{Look and Keep}
Next, all hypotheses $h_i \in \mathcal{H}$ are verified on the full image set $\mathcal{D}$ and the concept graph updated accordingly.  

\paragraph{Verification.} 
\texttt{ImageSet2Text} evaluates each hypothesis \( h_i \) $ \in \mathcal{H}$ against the entire image set $\mathcal{D}$ by leveraging the zero-shot classification capabilities of a CVL, following a one-vs-all classification problem, where positive and negative examples are generated for a given hypothesis \( h_i \), drawing from $\mathcal{G}_l$. Let $\mathcal{H}_i^+$ denote the set of other hypotheses that support \( h_i \), which are constructed by substituting the object $o_i$ with its hyponyms in $\mathcal{G}_l$; and let $\mathcal{H}_i^-$ denote the set of hypotheses that contradict \( h_i \), which are constructed by substituting $o_i$ with its sibling nodes in $\mathcal{G}_l$. Note that the supporting set $\mathcal{H}_i^+$ is expanded to include \( h_i \) itself. In the given example, for \( h_1 = \langle \text{`\textit{desert}'}, \text{`\textit{color}'}, \text{`\textit{yellow}'} \rangle \), supporting hypotheses replace `\textit{yellow}' with its hyponyms $o_h$, while contradicting ones use sibling nodes $o_s$:

\begin{tcolorbox}[colback=desertlavender, colframe=black!70, boxrule=1pt, arc=1mm, left=0pt]

$\mathcal{H}_1^+ = \{{ \langle \text{\mytiny{`\textit{desert}'}}, \text{\mytiny{`\textit{color}'}}, o_h \rangle \; \forall o_h \in \{\text{\mytiny{`\textit{gold}', `\textit{gamboge}', ...}}\}}\}$
$\mathcal{H}_1^- = \{{ \langle \text{\mytiny{`\textit{desert}'}}, \text{\mytiny{`\textit{color}'}}, o_s \rangle \; \forall o_s \in \{\text{\mytiny{`\textit{red}', `\textit{orange}', ...}}\}}\}$
\end{tcolorbox}

Next, all images in $\mathcal{D}$, along with each $o_h$ and $o_s$ are projected one-by-one into the CVL latent space and L2-normalized, yielding the sets of embeddings $\mathcal{E}_{\mathcal{D}}$, $\mathcal{E}_{\mathcal{H}_i^+}$, and $\mathcal{E}_{\mathcal{H}_i^-}$, respectively. A weighted \( k \)-Nearest Neighbors (kNN) classifier is then applied using the embeddings from \( \mathcal{E}_{\mathcal{H}_i^{+}} \) as positive examples and those from \( \mathcal{E}_{\mathcal{H}_i^{-}} \) as negative examples, with cosine similarity serving as the weighting metric. In particular, the kNN classifies each image \( x_j \in \mathcal{D} \) as supporting the hypothesis \( h_i \) if its corresponding embedding \( e_j \in \mathcal{E}_{\mathcal{D}} \) is labeled as positive, and as contradicting $h_i$ otherwise. As a result, the hypothesis $h_i$ is rejected if it is not verified on at least a predefined minimum portion $\alpha$ of the images in $\mathcal{D}$. Since hypotheses follow a hierarchical structure, if a hypothesis \( h_i \) is rejected, then any more specific hypothesis \( h_{i-1} \subset h_i \) is also unverified. This follows from the logical implication that $h_{i} \Rightarrow h_{i+1}$.

The hypotheses are verified in a general-to-specific manner to ensure semantic consistency in the CVL embedding space. If a highly specific hypothesis is prematurely tested without confirming its general category first, there is a risk of making comparisons in an embedding subspace that is not reliable \cite{radford2021}. Similarly, the set $\mathcal{H}_i^-$ is used because computing cosine similarities solely within $\mathcal{H}_i^+$ would not provide a reliable basis for evaluating the validity of a hypothesis. Comparisons against negative examples are necessary to establish a meaningful criterion for hypothesis acceptance \cite{chattopadhyay2024}. 

\paragraph{Concept Graph update.} At the end of the verification process, let \( h_\checkmark = \langle s, p, o_\checkmark \rangle \) be the most specific hypothesis \( h_i \in \mathcal{H} \) that is verified;  \( h_\checkmark \) is appended to the graph \( \mathcal{G}_c^{\tau+1} \) and the list of still‑pending predicates $P$ from $o_0$ is carried over to the next iteration (as illustrated in \cref{fig:sketch}). The object $o_\checkmark$, now a leaf node in \( \mathcal{G}_c^{\tau+1} \), is thus eligible to become the subject $s$ in a next iteration.

\subsection{Stopping Conditions}

An iteration in \texttt{ImageSet2Text} interrupts if any of the following conditions occurs: \textbf{1)} the VQA module flags a question as invalid (\eg, unsafe, inappropriate, unrelated to the content of the image) for at least a predefined number of images \( \theta \) in $\mathcal{S}$; \textbf{2)} no hypothesis in \(\mathcal{H}\) is verified for \(\mathcal{D}\); \textbf{3)} the updated graph \( \mathcal{G}_c^{\tau+1} \) adds no new information when compared to $\mathcal{G}_c^{\tau}$ as per an LLM evaluation. The entire iterative process ends when: \textbf{1)} no further graph expansion is possible, \ie, all existing nodes in \(\mathcal{G}_c^\tau\) have been explored; or \textbf{2)} a certain number \(\epsilon\) of consecutive iterations are discarded according to the previously mentioned criteria. Once the iterative process ends, any pending predicate in $P$ is discarded, and the final textual description \(d\) is generated directly from \(\mathcal{G}_c=\mathcal{G}_c^T\) using the LLM, as illustrated in \cref{fig:sketch}. 

\section{Evaluation}
\label{sec:evaluation}

We report the results of evaluating the generated descriptions (qualitative examples available in App. \ref{apx:sec:accuracy_baselines}) according to three properties, as depicted in Fig. \ref{fig:evaluation_pipeline}. 

\paragraph{Implementation details}
We use GPT-4o-mini \cite{achiam2023} as the LLM (including for VQA), Open-CLIP ViT-bigG-14 \cite{ilharco2021} as the CVL, and WordNet \cite{miller1995} as the external lexical graph. Each random subset $\cal{S}$ contains $M=10$ images; hypotheses are rejected if their verification rate falls below $\alpha = 0.8$ with $k=1$; iterations terminate upon encountering $\theta=10$ invalid images; traversing the knowledge hierarchy is limited to $\delta=2$ steps; and the maximum consecutive discarded iterations for stopping is $\epsilon=5$. Further details are provided in App.~\ref{sec:method_supp}.

\subsection{Accuracy}
\label{subsec:accuracy}

Accuracy is evaluated through the task of group image captioning. This task tests whether models can identify and describe common visual elements in a group of images, making it an ideal proxy for this property of the descriptions. 

\paragraph{Datasets} We did not find publicly available benchmarks for group image captioning of either small (up to 30 images) or large (up to thousands of images) image sets. Hence, we created two new benchmark datasets for this part of the evaluation (details in App.~\ref{apx:sec:accuracy:dataset}): 

\begin{itemize}
    \item \textbf{GroupConceptualCaptions}: Images with the same caption from the Conceptual Captions dataset \citep{sharma2018} are grouped. Each group's caption serves as ground truth, providing concise single-sentence descriptions of the main visual content. This dataset contains $116$ groups with a total of $23,412$ images.

    \item \textbf{GroupWikiArt}: WikiArt artworks \citep{artgan2018} with identical style, genre, and artist, are grouped, and the group captions are derived from these attributes. This dataset allows to evaluate nuanced, abstract concepts emerging from shared artistic interpretations within groups, and requires examining multiple images to capture group similarities. The dataset contains $105$ groups and $53,707$ images.
\end{itemize}

\paragraph{Models}
Given the lack of public models for group image captioning, we compare \texttt{ImageSet2Text} with four state-of-the-art vision-language models: BLIP-2 \cite{li2023}, LLaVA-1.5 \cite{liu2023}, GPT-4o \cite{achiam2023}, and Qwen2.5-VL \cite{bai:2025}. As these are designed for single-image captioning, we adapt them to group captioning through three strategies: (1) prompting image grids of varying sizes in a single image, (2) averaging image embeddings before caption generation (for open-source models), and (3) summarizing individual  captions into a group caption using GPT-4o. Further details are available in App.~\ref{apx:sec:accuracy_baselines} and App.~\ref{apx:sec:acc_compute}. Tests on multi-image settings are reported in App.~\ref{apx:sec:multi_image}. For \texttt{ImageSet2Text}, we generate captions by summarizing the full description into a single short sentence using GPT-4o, as detailed in App.~\ref{apx:sec:captionizing}. 

\paragraph{Metrics} Accuracy is quantified by measuring the semantic alignment between generated descriptions and ground-truth captions on both GroupConceptualCaptions and GroupWikiArt. Results are reported as the average rank across seven standard captioning metrics: four model-free,\footnote{CIDEr-D, SPICE, METEOR, ROUGE-L.} two model-based,\footnote{BERTScore F1, LLM-as-a-judge.} and one reference-free.\footnote{CLIPScore.} Full evaluation details can be found in App.~\ref{apx:sec:accuracy_metrics} and \ref{apx:sec:detailed_results}.

\begin{figure}[t]
    \centering
    \includegraphics[width=0.9\linewidth, trim=0.1cm 0.3cm 0.1cm 0, clip]{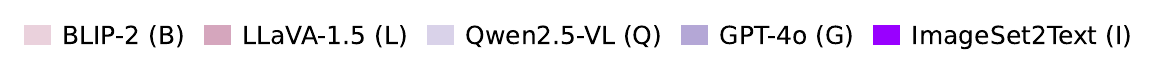}
    \begin{subfigure}[t]{0.5\linewidth}
        \centering
        \includegraphics[width=0.9\linewidth, trim=0.1cm 0.4cm 0.1cm 0, clip]{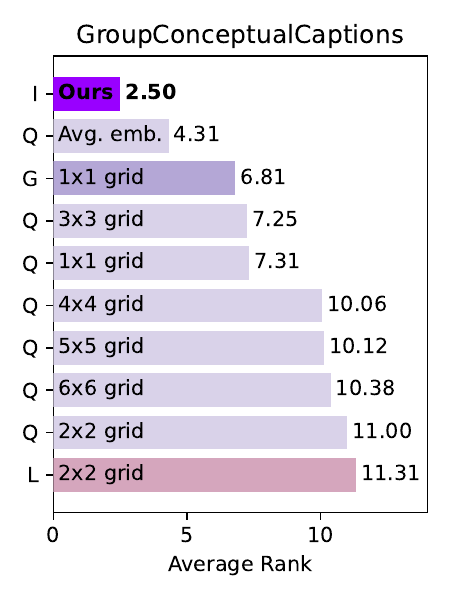}
    \end{subfigure}%
    \hfill
    \begin{subfigure}[t]{0.5\linewidth}
        \centering
        \includegraphics[width=0.9\linewidth, trim=0.1cm 0.4cm 0.1cm 0, clip]{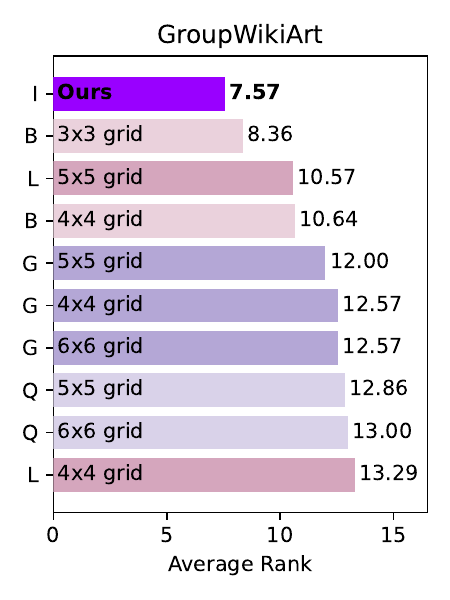}
    \end{subfigure}
    \caption{Accuracy as average rank across seven metrics on GroupConceptualCaptions (left) and GroupWikiArt (right). Only top ten methods shown; full results in App.~\ref{apx:sec:detailed_results}. 
    }
    \label{fig:accuracy_results}
\end{figure}

\paragraph{Results}

As shown in \cref{fig:accuracy_results}, \texttt{ImageSet2Text} achieves the \textbf{best} performance on both GroupConceptualCaptions and GroupWikiArt, with average ranks of $2.50$ and $7.57$, respectively, outperforming all baselines. 
In contrast, the performance of the baselines is not consistent across datasets: while Qwen2.5-VL is the second-best model in  GroupConceptualCaptions, BLIP-2, LLaVA-1.5, and GPT-4o exhibit stronger results in GroupWikiArt. Metric-wise analysis (detailed in App.~\ref{apx:sec:detailed_results}) reveals that \texttt{ImageSet2Text} performs particularly strong on model-based and reference-free metrics, which better capture semantic content and align more closely with human judgment \cite{zhang2020}.

\subsection{Completeness} 
\label{subsec:complete}
Completeness is evaluated by means of the Set Difference Captioning (SDC) task, which consists of identifying differences between two image sets, $\mathcal{D}_A$ and $\mathcal{D}_B$. To perform this comparison, we rely on the concept graphs generated by \texttt{ImageSet2Text} for each set. If the concepts are incomplete or vague, they would lead to failures in detecting differences between the sets. Hence, SDC serves as an effective proxy for evaluating the completeness of the underlying information used to generate the final descriptions---and, by extension, the completeness of the descriptions themselves.

\paragraph{Dataset} We adopt the methodology and dataset (PIS) described in \cite{visdiff}. The PIS dataset consists of $150$ contrastive set pairs (with $30,000$ images in total) spanning easy, medium, and hard distinctions. Consistent success in this task across varying difficult levels indicates that the description captures a range of relevant details necessary to differentiate $\mathcal{D}_A$ from $\mathcal{D}_B$.

\paragraph{Models} The PIS dataset was introduced together with VisDiff \cite{visdiff}, a proposer-ranker framework in which an LLM-based proposer suggests potential differences between image sets, and a ranker evaluates and ranks them using CLIP embeddings. The proposer relies on captions created with BLIP-2 from subsets of the two original datasets to identify differences between the two image sets. We compare VisDiff with \texttt{ImageSet2Text} by replacing the BLIP-2 captions with the concept graph representations produced by \texttt{ImageSet2Text}, keeping the rest of the framework unchanged (details in App. \ref{apx:graph2visdiff} and \ref{apx:com:compute}). 

\paragraph{Metrics} We use standard metrics: acc@1 and acc@5.

\paragraph{Results} Results are shown in \cref{tab:visdiff}. Using \texttt{ImageSet2Text} improves the performance with respect to the original VisDiff in all metrics except for acc@5 on the easy sets, where both methods achieve near-perfect accuracy ($0.99$). Importantly, \texttt{ImageSet2Text} consistently improves accuracy on the medium and hard sets, showing that its richer semantic representations enable better identification of subtle differences between the image sets (\eg, distinguishing between cars with metallic paint and matte paint). Indeed, the lack of details in the BLIP-2 captions lead to a set of failure cases in VisDiff that are mostly addressed with \texttt{ImageSet2Text} (see App.~\ref{apx:com:fails}). 

\renewcommand{\arraystretch}{0.95} 
\setlength{\tabcolsep}{10pt} 
\begin{table}
\caption{Completeness evaluation on the PIS dataset. The best performance per difficulty category is shown in bold.}
\label{tab:visdiff}
\centering
\begin{small} 
\begin{tabular}{llcc}
\toprule
Method & Category & Acc@1 & Acc@5 \\
\midrule
VisDiff  &  Easy & 0.88 & \textbf{0.99}  \\
 & Medium & 0.75 & 0.86 \\
 & Hard & 0.61 & 0.80 \\
\midrule
\texttt{ImageSet2Text} & Easy &  $\textbf{0.90}$ & $\textbf{0.99}$\\
& Medium & $\textbf{0.77}$ & $\textbf{0.89}$\\
& Hard & $\textbf{0.66}$ & $\textbf{0.82}$ \\ 

\bottomrule
\end{tabular}
\end{small}
\end{table}
\renewcommand{\arraystretch}{1}

\subsection{User Satisfaction}
While the previous experiments measure accuracy and completeness, they do not assess the overall quality of the descriptions, which we evaluate in a user study conducted on a random subset of the PIS dataset with $233$ users.

\paragraph{Methodology}

We sample 60 PIS image sets (20 per difficulty level) and, for each, we show participants 16 images in a $4 \times 4$ grid alongside a description (see App.~\ref{sec:user_supp}). Participants are asked to rate the \emph{clarity, accuracy, detail, flow,} and \emph{overall satisfaction} of the descriptions on a five $5$-point Likert-scale. To ease the interpretation of the user feedback and given that there is a lack of alternative methods to create descriptions of large image sets, we generate $10$ control descriptions using ChatGPT (see App.~\ref{sec:user_supp} for examples), divided into three categories: 

\begin{itemize}
    \item \textbf{Control accuracy} (3 descriptions): well-written but intentionally inaccurate descriptions.
    \item \textbf{Control detail} (3 descriptions): descriptions that reference the correct visual content but lack details.
    \item \textbf{Control clarity and flow} (4 descriptions): factually correct, detailed but with low coherence descriptions.
\end{itemize}

Note that comparing against the captions described in \cref{subsec:accuracy} would be inappropriate due to length and detail mismatches that could bias results \cite{grice1975logic,kahneman2011thinking}. Also note that the control descriptions are not performance baselines: they isolate specific qualities, allowing for controlled comparisons in alignment with human-centered evaluation best practices \cite{rogers2023interaction}.

A total of $233$ qualifying participants\footnote{18+ native English-speakers w/o visual/reading impairments.} were recruited via Prolific,\footnote{\url{https://www.prolific.com/} (accessed 2025/03/07).} each evaluating $7$ descriptions ($6$ \texttt{ImageSet2Text}, $1$ control) and performing $2$ attention tests. $198$ participants successfully completed the study, yielding $16-22$ evaluations per description. The average task completion time was $8$ minutes, compensated at \$12/hour. The process was fully anonymized, and no personal information was collected.   

\paragraph{Results}

\begin{figure}[t]
    \centering
    \includegraphics[width=0.9\linewidth]{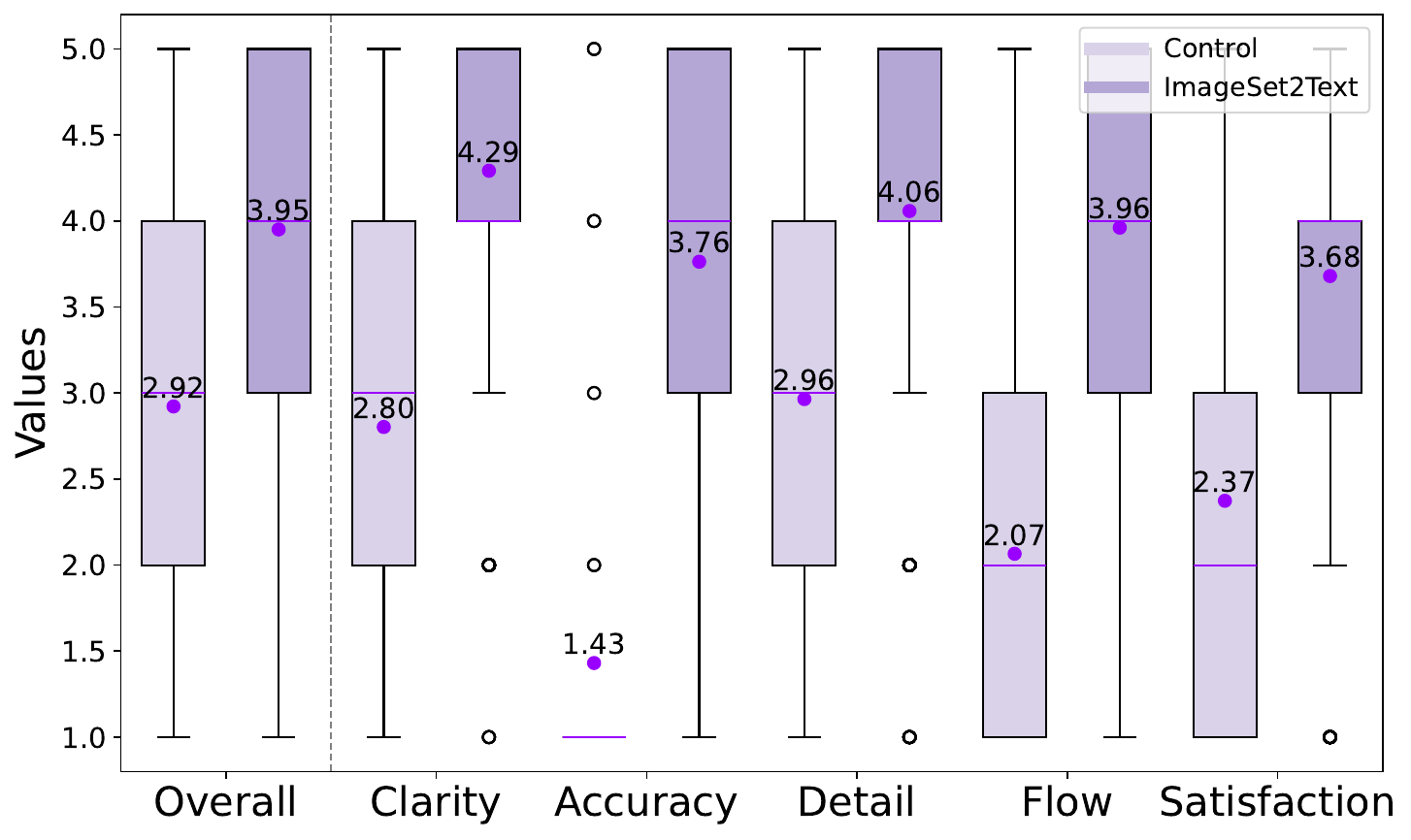}
    \caption{User study results. For control, values are of those designed to assess clarity, accuracy, detail, and flow, respectively, whereas overall and satisfaction are averaged across all control descriptions. Purple bars indicate medians, dots show means with the actual values reported in the figure.}
    \label{fig:user_study}
\end{figure}

Results are shown in \cref{fig:user_study}. \texttt{ImageSet2Text} descriptions are consistently rated favorably, not only in absolute terms but also when compared to the reference levels established by the control descriptions: clarity ($\mu=4.29$ vs control $\mu=2.80$), accuracy ($\mu=3.76$ vs control $\mu=1.43$), detail ($\mu=4.06$ vs control $\mu=2.96$) and flow ($\mu=3.96$ vs control $\mu=2.07$). These differences are statistically significant (t-test, all $p \ll 10^{-5}$). 

\section{Method's behavior}
\label{sec:ablation} 
We report analyses on \texttt{ImageSet2Text}'s behavior, considering an ablation study, scalability estimate and failure cases analysis (Fig. \ref{fig:evaluation_pipeline}) and discussing potential applications.

\paragraph{Ablation study}

Inspired by ongoing research that combines symbolic and data-driven AI \cite{marcus2018, guo2024}, we conducted an ablation study with four versions of \texttt{ImageSet2Text}, each progressively integrating more structured information:

\begin{enumerate}[label=\texttt{v\arabic*}]
\item relies only on LLMs to generate hypothesis \(h_0\), the set of general hypotheses \(\mathcal{H}\), and the supporting \(\mathcal{H}_i^+\) and contradicting \(\mathcal{H}_i^-\) alternatives. The extracted insights are directly used to refine the description iteratively; 

\item  introduces the external lexical graph $\mathcal{G}_l$ to generate the sets \( \mathcal{H}\), \(\mathcal{H}_i^+\) and \(\mathcal{H}_i^-\). No concept graph is kept in memory and the textual description is updated at every iteration; 
\item introduces the iterative concept graph \(\mathcal{G}_c^\tau\), with the final description being generated from  \(\mathcal{G}_c^T\). 
This is the version of \texttt{ImageSet2Text} introduced in this paper; 
\item decomposes hypothesis formulation into two steps: (1) the LLM to summarize the VQA answers $\mathcal{A}$ into a sentence, and (2) POS tagging and dependency parsing to extract the object $o_0$ and its candidate predicates. 
\end{enumerate}

\setlength{\tabcolsep}{1pt} 
\begin{table}
\caption{Ablation study with incremental structured knowledge representation on a subset of the PIS dataset (* indicates the version corresponding to \texttt{ImageSet2Text}).}
\label{tab:ablation}
\centering
\begin{small} 
\begin{tabularx}{\columnwidth}{lccccrr}
\toprule
& \footnotesize{\quad $\mathcal{G}_l$ \quad} & \footnotesize{\quad $\mathcal{G}_c^\tau$ \quad} & \footnotesize{\quad dependency parsing \quad} & \footnotesize{Acc@1} & \footnotesize{Acc@5} \\
\midrule
\texttt{v1}  & - & - & - &  0.67 & 0.87\\
\texttt{v2}  & \checkmark & - & - & 0.77 & 0.87\\
\texttt{v3}*  & \checkmark & \checkmark & - & \textbf{0.90} & \textbf{1.00}\\
\texttt{v4}  & \checkmark & \checkmark  & \checkmark & 0.67 & 0.87\\
\bottomrule
\end{tabularx}
\end{small}
\vskip -0.1in
\end{table}

We report completeness (as in \cref{subsec:complete}) on a random subset of $15$ image set pairs ($5$ easy, $5$ medium, and $5$ hard).
\cref{tab:ablation} shows that the progressive integration of structured information improves the performance up to \texttt{v3}, followed by a decline in \texttt{v4}. This result is confirmed by manual assessment, where \texttt{v3} produced the highest quality descriptions. The generation of \( \mathcal{H}\),  \(\mathcal{H}_i^+\), and \(\mathcal{H}_i^-\) in \texttt{v2} is more effective compared to \texttt{v1}, which is subject to hallucinations due to relying only on the LLM. However, both \texttt{v2} and \texttt{v1} occasionally produce descriptions with a broken flow. This limitation is addressed in \texttt{v3} by directly generating the final description from the concept graph. POS tagging and dependency parsing in \texttt{v4} are error-prone and hard to adapt, making them less reliable than the LLM alone for hypothesis formulation. In conclusion, the best-performing version, \texttt{v3}, is the version that best leverages the advantages of both symbolic and data-centric AI. 

\paragraph{Scalability}
\texttt{ImageSet2Text} scales efficiently to large image sets of size $N$. Hypotheses are generated from a small random sample $M \ll N$, keeping computational cost independent of $N$ (with $M$ depending on set heterogeneity rather than size). For verification, $1280$-dimensional embeddings (\ie, $d = 1280$) for all $N$ images are precomputed \textit{once} on the CVL model at $\approx$ $12$ images/s on an NVIDIA RTX 3090, requiring $T_{embed}(N) = N/12$ seconds.\footnote{\url{https://gist.github.com/TACIXAT/ecd4f636bf6af28cb69d641e29d7b362} (accessed 2025/07/28)} The kNN step, repeated over $T$ iterations, compares $N$ images with $S+C$ example points, taking $T_{iter}(N, |\mathcal{H}_i^+|, |\mathcal{H}_i^-|) = N \cdot (|\mathcal{H}_i^+|+|\mathcal{H}_i^-|) \cdot d \cdot 2/ \mathrm{FLOPs}$ seconds, where $\mathrm{FLOPs}$ denotes the GPU’s floating-point throughput. The factor of 2 accounts for one multiplication and one addition per dimension when computing cosine distance on L2-normalized embeddings. For example, using FP32 precision on an NVIDIA RTX 3090 (35.58 TFLOPs), with $1$ million images and $|\mathcal{H}_i^+|+|\mathcal{H}_i^-| = 1000$, each kNN iteration takes $<0.1$~s, which is \texttt{ImageSet2Text}’s only $N$-dependent step. Across our experiments, $T$ ranges from $10$ to $30$, independently of $N$. In addition, avg graphs $\#$nodes (depth) are: 10.84 (3.28) in GroupConceptualCaptions, 9.83 (3.13) in GroupWikiArt; in PIS: 10.84 (3.31) easy, 10.48 (3.48) medium, 10.37 (3.57) hard, indicating fewer nodes but similar/higher depth due to higher visual diversity.

\begin{figure}[t]
    \centering
    \includegraphics[width=\linewidth]{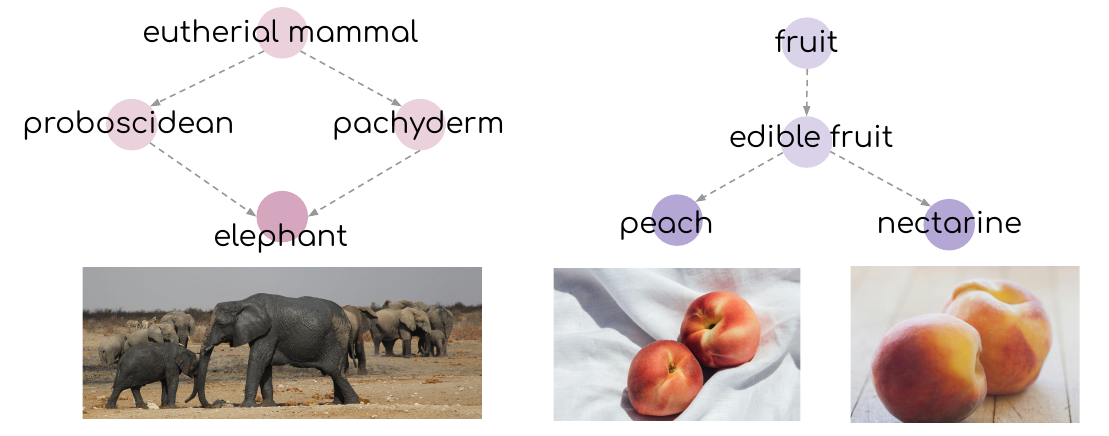}
    \caption{Illustrative examples of failure cases: (1) sibling nodes that are not mutually exclusive (\textit{pachyderm} vs.\ \textit{proboscidean}), causing incorrect formulation of $\mathcal{H}_i^-$; and (2) sibling nodes that cannot be visually distinguished (\textit{peach} vs. \textit{nectarine}), causing kNN misclassifications.}
    \label{fig:failures}
\end{figure}

\paragraph{Failure cases}

We identify two main failure types (see Fig. \ref{fig:failures}) due to the integration of WordNet and CLIP: (1) sibling nodes that are not mutually exclusive cause $\mathcal{H}_i^-$ elements to be supportive (in 1.49\% of cases); (2) sibling nodes that are hard to distinguish visually cause kNN misclassifications (in around 2\% of cases). While these issues affect the verification process, they are sufficiently rare to have minimal overall impact. Details on how we estimate their incidence are provided in App.~\ref{apx:failures}, along with other limitations inherited from the modules constituting \texttt{ImageSet2Text}.

\paragraph{Potential Applications}
The high readability of the descriptions generated by \texttt{ImageSet2Text} suggests broader applications beyond the already demonstrated possibilities of the SDC task \cite{visdiff} (Section 4.2), such as dataset exploration under user-guidance, explainable AI, cultural analytics, and the identification of potential biases in image sets. Additionally, we initiated a collaboration with Fundación ONCE\footnote{https://www.fundaciononce.es/en (accessed 2025/11/11)} to collect feedback on \texttt{ImageSet2Text}'s value for visually-impaired individuals, incorporating community insights early in the design process \cite{costanza2020}. Details in App.~\ref{sec:app_supp}.

\section{Conclusion}

In this paper, we have proposed \texttt{ImageSet2Text}, a novel method to generate natural language descriptions of image sets. We have shown its competitive performance in: (1) a large-scale group captioning experiment with two newly proposed benchmarks (GroupConceptualCaptions and GroupWikiArt); (2) set difference captioning to assess description completeness; and (3) a user study. Through an ablation study, a scalability analysis, and failure case examination, we have illustrated how \texttt{ImageSet2Text} successfully integrates symbolic and data-centric approaches. Overall, \texttt{ImageSet2Text} effectively describes large image collections, proving valuable for diverse applications.

\section*{Acknowledgements} 
We are grateful to Fundación ONCE for their willingness to collaborate on our research. \textbf{PR} and \textbf{NO} have been partially supported by a nominal grant received at the ELLIS Unit Alicante Foundation from the Regional Government of Valencia in Spain (Resolución de la Conselleria de Innovación, Industria, Comercio y Turismo, Dirección General
de Innovación). \textbf{PR }was also supported by a grant by Fundación Banc Sabadell. \textbf{NG} is partly supported by JSPS KAKENHI No.
JP22K12091. \textbf{KS} is part of the ELLIS Unit Linz, the LIT AI Lab and the Institute for Machine Learning at Johannes Kepler University Linz, which are supported by the Federal State Upper Austria. We thank the projects FWF AIRI FG 9-N (10.55776/FG9), AI4GreenHeatingGrids (FFG- 899943), Stars4Waters (HORIZON-CL6-2021-CLIMATE-01-01), FWF Bilateral Artificial Intelligence (10.55776/COE12). We thank NXAI GmbH, Audi AG, Silicon Austria Labs (SAL), Merck Healthcare KGaA, GLS (Univ. Waterloo), T\"{U}V Holding GmbH, Software Competence Center Hagenberg GmbH, dSPACE GmbH, TRUMPF SE + Co. KG. We thank \textbf{Federico Brunero}, \textbf{Julien Colin}, \textbf{Erik Derner}, \textbf{Aditya Gulati}, \textbf{Benedikt Höltgen}, \textbf{Fabian Paischer} and \textbf{Korbinian Pöppel} for helpful discussions.

{\small
\bibliography{aaai2026}
}

\clearpage

\setcounter{secnumdepth}{2}
\appendix
\section{ImageSet2Text: Implementation Details}
\label{sec:method_supp}

In this section, we provide details regarding the prompts used for the LLM in our implementation of \texttt{ImageSet2Text}. The code will be publicly shared through our GitHub repository.

To effectively prompt GPT-4o-mini, we design system prompts that outline the model’s role and guide its reasoning process. Additionally, we utilize structured outputs to ensure consistency in responses. For example, we reference the ``next question" prompt used in \texttt{ImageSet2Text}, provided in \cref{fig:next_question_combined}, to explain the rationale behind our prompt engineering. This is a key component in our design, as it provides a structured method for generating follow-up questions to refine image set descriptions.
  
The system prompt defines GPT-4o-mini’s role as an expert assisting in generating textual descriptions for a large image set.  Specifically, the model is given:  

\noindent \textbf{QUESTION\_BRANCH}: A connection within our graph representation which highlights an aspect of the image set that requires further exploration.  

\noindent \textbf{KEY\_POINT}: The subject at the current iteration, \emph{i.e.}, the main element within the description that needs additional details. 

\noindent \textbf{ATTRIBUTE}: The predicate at the current iteration, \emph{i.e.}, the specific property of the \textbf{KEY\_POINT} to investigate further.  

\noindent \textbf{LOG}: A history of previously asked questions about the \textbf{KEY\_POINT} to avoid redundancy.
  
To ensure coherence and usability, the model produces responses in JSON format with two key fields:  

\noindent \textbf{QUESTION\_EXPERT}: A refined, expert-level question that directly addresses the given \textbf{ATTRIBUTE} in a way that enriches the overall understanding of the image set.  

\noindent \textbf{QUESTION\_VQA}: A simplified, image-focused translation of \textbf{QUESTION\_EXPERT} that adheres to the constraints of a Visual Question Answering (VQA) system. The design of \textbf{QUESTION\_VQA} follows strict guidelines:  
    \begin{itemize}  
        \item It must be direct, clear, and reference visible elements in a single image.  
        \item It should avoid abstract reasoning, cultural knowledge, or domain expertise beyond visual interpretation.  
        \item It must elicit descriptive responses rather than simple yes/no answers.  
        \item It should ensure novelty, avoiding redundancy with previously asked questions.  
    \end{itemize}   

This structured approach overall improves the quality of GPT-4o-mini. The other prompts used for the LLM in \texttt{ImageSet2Text} follow the same criteria. As a further example, in Figure \ref{fig:graph2text} we also provide the graph-to-text rendering prompt, used at the final stage of the pipeline of \texttt{ImageSet2Text}.

\begin{figure*}[]
    \centering
    \begin{subfigure}[b]{0.8\textwidth}
        \lstset{
            basicstyle=\tiny\ttfamily, 
            frame=single, 
            breaklines=true, 
            breakatwhitespace=true
        }
        \begin{lstlisting}
NEXT_QUESTION = """You are an expert in {expertise} assisting a client in enriching the textual description of a large IMAGE_SET in their possession.

Using JSON, you will be provided with a question expressed as a connection within a knowledge graph built for IMAGE_SET (QUESTION_BRANCH), a KEY_POINT that lacks detailed explanation in DESCRIPTION_CURRENT, an ATTRIBUTE which is the specific property of the KEY_POINT to investigate further, and a list of textual questions which have already been asked about that KEY_POINT (LOG).

An example of a client query might be the following:

{jsonScheme_input}

Based on this input, translate the QUESTION_BRANCH into two textual concise questions (QUESTION_EXPERT and QUESTION_VQA) to investigate the ATTRIBUTE further.

Provide the answer in JSON format. For example, the output from an expert in couples would be:

{jsonScheme_output}

In the output, it is expected that:
- The field QUESTION_EXPERT is a string containing an expert-level query that specifically addresses the selected ATTRIBUTE, intended to enrich the understanding of the IMAGE_SET.
- The field QUESTION_VQA is a string containing the translation of QUESTION_EXPERT into an image-focused question that meets the following criteria:
  1. QUESTION_VQA must be direct, simple, clear, and must refer to visible aspects within the image.
  2. QUESTION_VQA should be structured to ask about details that can be observed in one randomly selected image from the IMAGE_SET, referring to "the image" directly, like "What is in the image?".
  3. Avoid questions that require abstract reasoning, cultural knowledge, or expertise beyond what can be visually interpreted.
  4. Avoid yes/no questions and focus instead on generating descriptive responses.
  5. Ensure that QUESTION_VQA aligns with the capabilities of a VQA system, such as object recognition, spatial relationships, counting, and scene understanding.
  6. QUESTION_VQA must be designed to elicit new insights without overlapping with previous responses.
        \end{lstlisting}
        \caption{``Next Question" Prompt}
        \label{lst:next_prompt}
    \end{subfigure}

    \vspace{1em} 
    \begin{subfigure}[b]{0.8\textwidth}
        \lstset{
            language=Python,
            basicstyle=\tiny\ttfamily, 
            frame=single, 
            breaklines=true, 
            breakatwhitespace=true
        }
        \begin{lstlisting}
def next_question(expertise):
    expertise = expertise_to_string(expertise)
    
    jsonScheme_input = """{
        "QUESTION_BRANCH": "image.wedding.couple.body language?",
        "KEY_POINT": "couple",
        "ATTRIBUTE": "body language: communication via the movements or attitudes of the body",
        "log": [
            {
                "ATTRIBUTE": "gender composition: the properties that distinguish organisms on the basis of their reproductive roles",
                "QUESTION_EXPERT": "What is the gender composition of the couple portrayed in the image?",
                "QUESTION_VQA": "How many men and women are visible in the image?"
            },
            {
                "ATTRIBUTE": "attire: clothing of a distinctive style or for a particular occasion",
                "QUESTION_EXPERT": "What is the attire of the couple portrayed in the image?",
                "QUESTION_VQA": "What type of clothing are the individuals in the image wearing?"
            }
        ]
    }"""

    jsonScheme_output = """{
        "QUESTION_EXPERT": "How does the couple's body language appear in the wedding image?",
        "QUESTION_VQA": "What are the body positions or movements of the spouses in the image?"
    }"""

    return NEXT_QUESTION.format(
        expertise=expertise, 
        jsonScheme_input=jsonScheme_input, 
        jsonScheme_output=jsonScheme_output
    )
        \end{lstlisting}
        \caption{Function for ``Next Question"}
        \label{lst:next_function}
    \end{subfigure}

    \caption{``Next Question" Prompt}
    \label{fig:next_question_combined}
\end{figure*}

\begin{figure*}[]
    \centering
    \begin{subfigure}[b]{0.8\textwidth}
        \lstset{
            basicstyle=\tiny\ttfamily, 
            frame=single, 
            breaklines=true, 
            breakatwhitespace=true
        }
        \begin{lstlisting}
WRITE_DESCRIPTION = """You are assisting a client in writing the textual description of a large IMAGE_SET in their possession.

The client has collected pieces of verified information about the IMAGE_SET in form of a knowledge graph, and has generated a network text representation out of it.  

Using JSON, you will be provided with the knowledge graph text representation (GRAPH).

An example of a client query might be the following:

{jsonScheme_input}

Given the GRAPH, write a textual description (DESCRIPTION) of the entire IMAGE_SET. Provide the answer in JSON format. For example:

{jsonScheme_output}

GUIDELINES
- As you write the DESCRIPTION, make sure to retain all existing details from the GRAPH. Do not miss any piece of information.
- Stick strictly to the details provided in the GRAPH without adding, assuming, or inventing any information.
- Note that all information found in the GRAPH applies to every image within the IMAGE_SET.
"""
        \end{lstlisting}
        \caption{``Graph to Text" Prompt}
        \label{lst:graph2text_prompt}
    \end{subfigure}

    \vspace{1em} 
    \begin{subfigure}[b]{0.8\textwidth}
        \lstset{
            language=Python, 
            basicstyle=\tiny\ttfamily, 
            frame=single, 
            breaklines=true, 
            breakatwhitespace=true
        }
        \begin{lstlisting}
def write_description():
    jsonScheme_input = "    {" + """
        "GRAPH": "-- ('entity', 'image')
    |-> ('image', 'content', 'mountains')
    |   |-> ('mountains', 'portrayed_during', 'golden hour')
    |-> ('image', 'type', 'photographs')"
    """ + "}"
    
    jsonScheme_output = "    {" + """
        "DESCRIPTION": "A collection of photographs capturing mountains, portrayed during the golden hour."
    """ + "}"

    return WRITE_DESCRIPTION.format(jsonScheme_input=jsonScheme_input, jsonScheme_output=jsonScheme_output)
        \end{lstlisting}
        \caption{Function for ``Next Question"}
        \label{lst:graph2text_function}
    \end{subfigure}

    \caption{``Graph 2 Text" Prompt}
    \label{fig:graph2text}
\end{figure*}

\section{Accuracy Evaluation}
\label{sec:accuracy_supp}

In this section, we provide additional information regarding the accuracy evaluation.
Specifically, we provide details about the dataset construction process, how to generate a caption with \texttt{ImageSet2Text}, details on how the baselines have been set up and the full results of the experiments.

\subsection{Dataset Composition and Construction} \label{apx:sec:accuracy:dataset}

\begin{figure}[htb!]
    \centering
    \includegraphics[width=\linewidth]{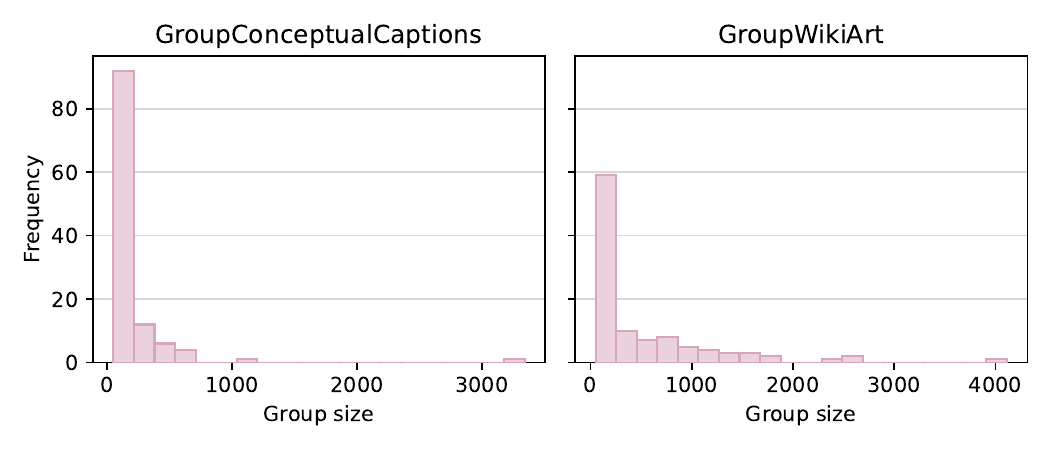}
    \caption{Group sizes for the GroupConceptualCaptions and GroupWikiArt datasets.}
    \label{fig:group_sizes}
\end{figure}

Given the absence of public benchmarks for evaluating \texttt{ImageSet2Text} in the group image captioning task, we construct two datasets.
Specifically, the GroupConceptualCaptions dataset based on the ConceptualCaptions dataset \citep{sharma2018} and the GroupWikiArt dataset based on the WikiArt dataset \cite{artgan2018}.
Group sizes for these datasets are depicted in \cref{fig:group_sizes}.
The minimal size is 50, and many groups are smaller than 1000 samples, yet there are also sets in the range of 3000-4000 images. To retrieve relevant image sets from each of the two data sources, we employed different techniques.

For the ConceptualCaptions dataset \cite{sharma2018}, we grouped images that shared the same caption and applied a filter to retain only those sets with more than 100 available links in the metadata of the original dataset. 
This process resulted in 287 distinct captions. 
After downloading the images, we found that not all links were accessible. 
We further filtered the sets, keeping only those with at least 50 images, which resulted in 125 unique sets. 
We then manually reviewed all the sets, removing 9 sets that contained either duplicate images, broken images, or images that did not correspond to the caption. 
Additionally, we deleted any duplicate images within the remaining 116 sets. As a result, we obtained 116 image sets, with sizes ranging from 50 to 3,342 images, for a total of 23,412 images.

The initial WikiArt dataset \cite{artgan2018} does not include explicit reference captions on which we could group the images, but it includes metadata about the artist, style, and genre. The possible values for these attributes are listed in \cref{tab:wikiart_values}. 
Using this information, we generated captions for the images based on the following rules: when aggregating by genre and style, the caption format is ``$<genre>$ in $<style>$ style''; when considering the artist as well, the caption format becomes ``$<genre>$ by $<artist>$ in $<style>$ style''. 
For instance, possible captions are: ``Landscapes in Romanticism Style'' or ``Religious paintings by Edgar Degas in Impressionism Style''. 
To ensure sufficient data for analysis, we filtered the groups based on the number of images: groups with only two attributes (style and genre) were kept if they contained more than 499 images, while groups with three attributes (style, genre and artist) required a minimum of 50 images. 
After downloading the images and removing duplicates, we retained only the groups with more than 49 images. 
From this process, we obtained 53,707 images distributed across 105 groups. 
The sizes of these groups range from 50 to 4,112 images.

\begin{table}[t]
\caption{Metadata values for creating group captions on WikiArt.}
\label{tab:wikiart_values}
\centering
\begin{small}
\begin{tabular}{p{\linewidth}}
\toprule
\multicolumn{1}{c}{\textbf{Artist}} \\
\midrule
0: ``Albrecht Durer'', 1: ``Boris Kustodiev'', 2: ``Camille Pissarro'', 3: ``Childe Hassam'',  
4: ``Claude Monet'', 5: ``Edgar Degas'', 6: ``Eugene Boudin'', 7: ``Gustave Dore'',  
8: ``Ilya Repin'', 9: ``Ivan Aivazovsky'', 10: ``Ivan Shishkin'', 11: ``John Singer Sargent'',  
12: ``Marc Chagall'', 13: ``Martiros Saryan'', 14: ``Nicholas Roerich'', 15: ``Pablo Picasso'',  
16: ``Paul Cezanne'', 17: ``Pierre Auguste Renoir'', 18: ``Pyotr Konchalovsky'', 19: ``Raphael Kirchner'',  
20: ``Rembrandt'', 21: ``Salvador Dali'', 22: ``Vincent van Gogh'' \\  
\midrule
\multicolumn{1}{c}{\textbf{Genre}} \\
\midrule
0: ``Abstract paintings'', 1: ``Cityscapes'', 2: ``Genre paintings'', 3: ``Illustrations'',  
4: ``Landscapes'', 5: ``Nude paintings'', 6: ``Portraits'', 7: ``Religious paintings'',  
8: ``Sketches and studies'', 9: ``Still lifes'' \\  
\midrule
\multicolumn{1}{c}{\textbf{Style}} \\
\midrule
0: ``Abstract Expressionism'', 1: ``Action painting'', 2: ``Analytical Cubism'', 3: ``Art Nouveau'',  
4: ``Baroque'', 5: ``Color Field Painting'', 6: ``Contemporary Realism'', 7: ``Cubism'',  
8: ``Early Renaissance'', 9: ``Expressionism'', 10: ``Fauvism'', 11: ``High Renaissance'',  
12: ``Impressionism'', 13: ``Mannerism Late Renaissance'', 14: ``Minimalism'', 15: ``Naive Art Primitivism'',  
16: ``New Realism'', 17: ``Northern Renaissance'', 18: ``Pointillism'', 19: ``Pop Art'',  
20: ``Post Impressionism'', 21: ``Realism'', 22: ``Rococo'', 23: ``Romanticism'', 24: ``Symbolism'',  
25: ``Synthetic Cubism'', 26: ``Ukiyo-e'' \\  
\bottomrule
\end{tabular}
\end{small}
\vskip -0.1in
\end{table}

\subsection{Baseline Methods} \label{apx:sec:accuracy_baselines}

\begin{figure}
    \centering
    \includegraphics[width=\linewidth]{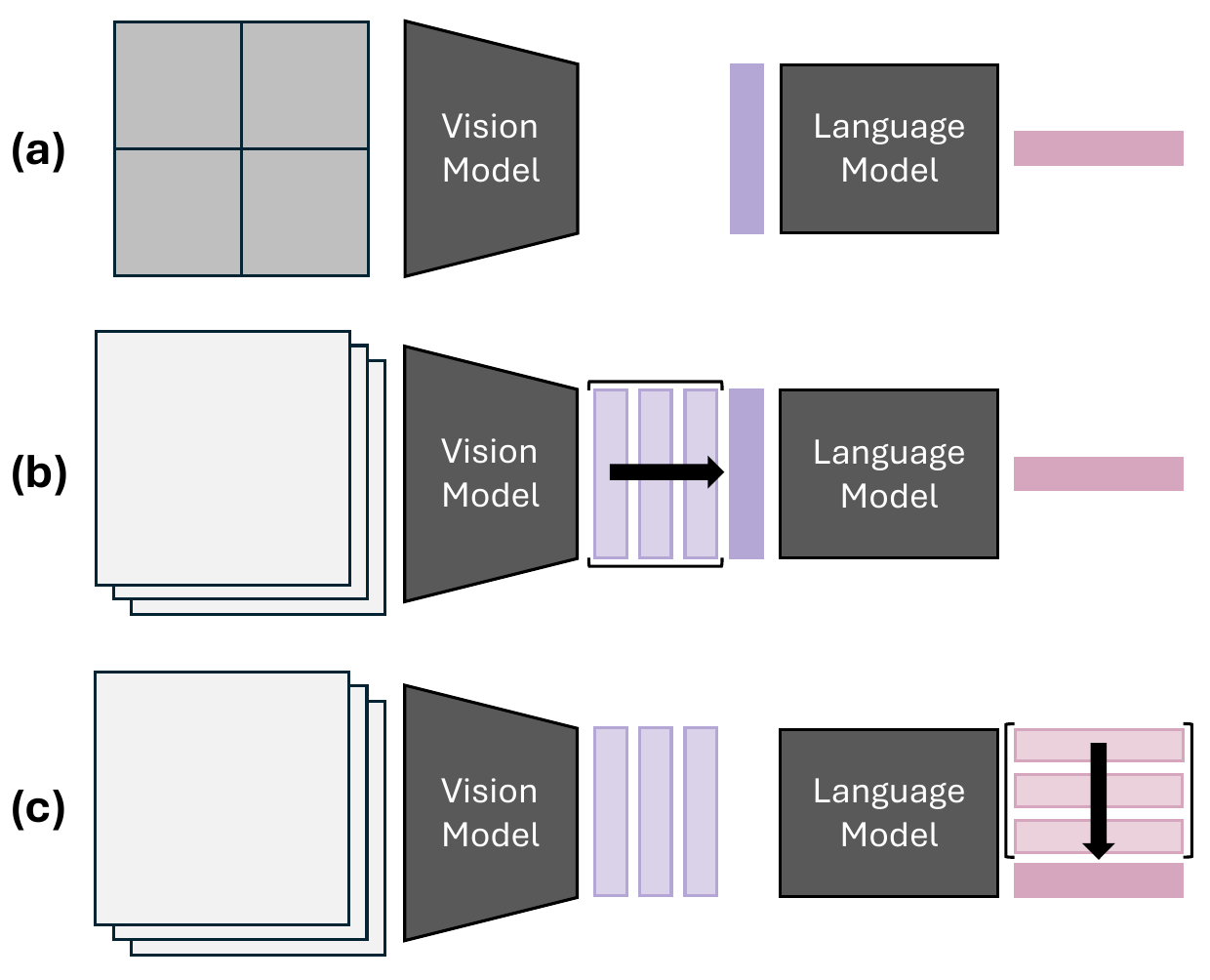}
    \caption{Settings of utilizing VQA models to generate group captions. Light colors denote images (gray), image embeddings (purple) and output text (pink) on the level of individual images, dark colors denote aggregation on a group level. (a) is the grid setting where images are put into a collage to depict the group, (b) is the group embedding setting where embeddings of individual images after the vision encoder are averaged and (c) is the summary setting where an additional LLM is used to generate a group caption from the captions of individual instances.}
    \label{fig:captioning_settings}
\end{figure}

The baselines we compare against for this task, namely BLIP-2 \cite{li2023}, LLaVA-1.5 \cite{liu2023}, GPT-4o \cite{achiam2023} and QWEN2.5-VL \cite{bai:2025}, are not designed to process multiple images simultaneously. 
Therefore, we conduct experiments using different settings to enable a comparison with \texttt{ImageSet2Text}. 
These settings are referred to as (a) the grid setting, (b) the group embedding setting, and (c) the summary setting. 
A visual summary of these settings is provided in \cref{fig:captioning_settings}.

In the grid setting, we select images from the groups, arrange them into grids, and use these grids as inputs for the baseline models. 
Due to resolution constraints on the input size of the respective vision encoders, we include only a subset of images from each group.
For the biggest groups of multiple thousand images, using the entire set would result in each image occupying only a few pixels within the grid, leading to meaningless outputs. 
To be applicable to all groups we investigated grid sizes of up to 7x7 images, which is close the minimum number of images present in the smallest groups (50).

In the group embedding setting, all images of a group are passed through the vision encoder of the respective model and are averaged before the caption is generated.
Naturally this is only applicable to open-source models, thus not to the GPT-4o baseline.

In the summary setting, we generate a caption for every individual image of the group.
Then we utilize GPT-4o to summarize these captions into a single caption.
We only considered unique captions.
The prompt format used for this operation is: \vspace{15pt}

\lstset{basicstyle=\scriptsize\ttfamily, frame=single, breaklines=true, breakatwhitespace=true}
\begin{lstlisting}
In the following, a list of captions is provided. 
Generate a caption that best describes the group of captions.
The group captions should be short and concise.
{[f"{i+1}: {caption} " for i, caption in enumerate(captions)]}
Group Caption:
\end{lstlisting}
\vspace{15pt}

To ensure a fair comparison with \texttt{ImageSet2Text}, the grid setting necessitates a ``caption curation'' step. 
Since the models process the grid as a single image, they often generate captions containing phrases like ``collage of images'' or ``grid of images''. 
These terms negatively impact performance evaluation, as the reference captions describe only the image content without mentioning grids or collages. 
Therefore, we remove such terms or sentences from the generated captions, evaluating only the detected content within the grid images.
To not introduce any systematic bias to the evaluation, we applied this curation step to captions of all methods, also \texttt{ImageSet2Text}.

\subsection{Generating a Caption with \texttt{ImageSet2Text}} \label{apx:sec:captionizing}

\begin{figure}
    \centering
    \includegraphics[width=\linewidth]{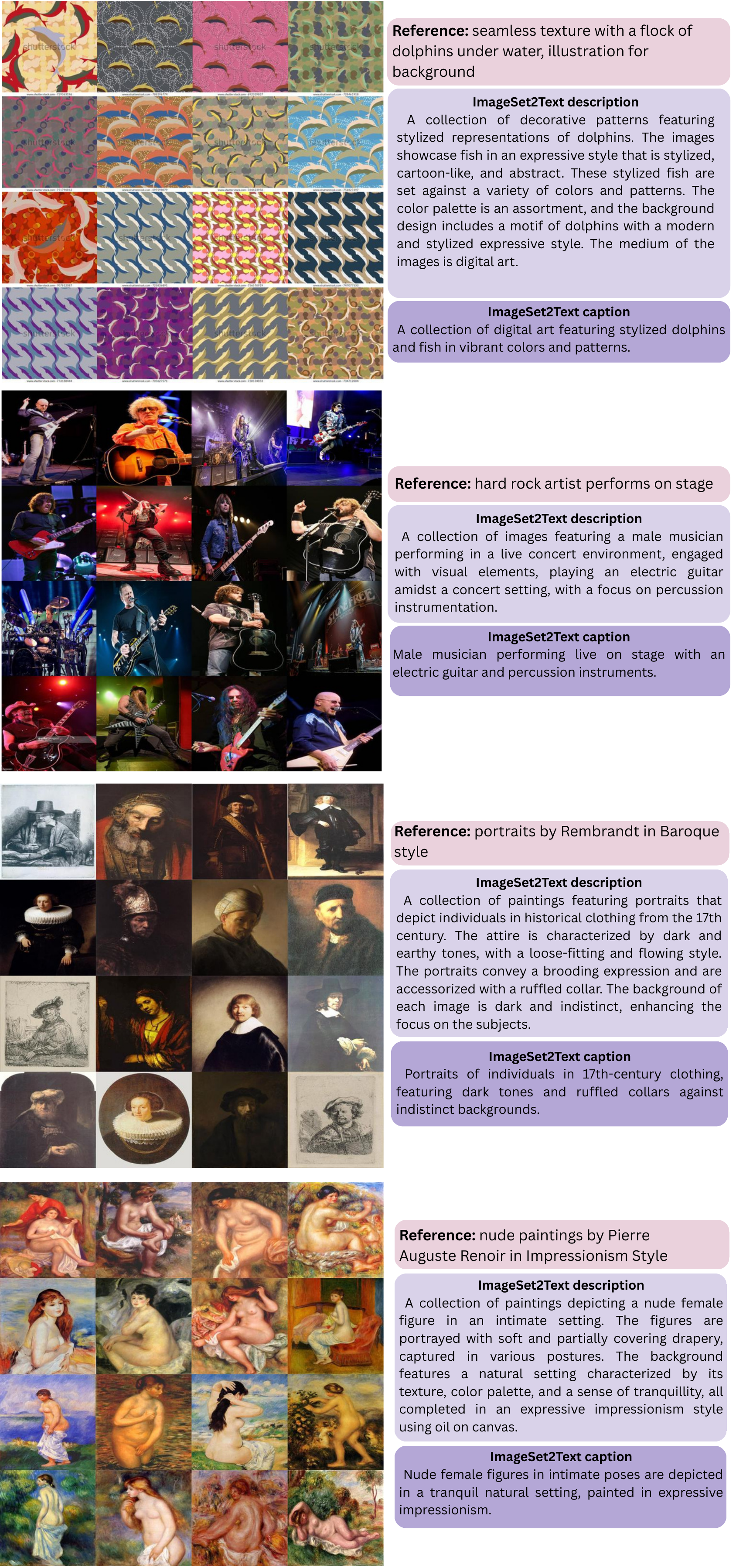}
    \caption{Examples of generated descriptions and generated captions for four different image sets. From top two bottom: two sets in GroupConceptualCaptions and two sets in GroupWikiArt.}
    \label{fig:gc}
\end{figure}

\texttt{ImageSet2Text} is not explicitly designed for group captioning, but this task serves as a key evaluation tool for assessing the accuracy of descriptions. 
The typical output of \texttt{ImageSet2Text} is a long, nuanced, and detailed textual description highlighting the main visual elements shared among the images in a given set. 
However, for evaluation purposes, this detailed description must be transformed into a more plain and concise textual representation, \emph{i.e.}, a caption. To generate a caption from a description issued by \texttt{ImageSet2Text}, we utilize GPT-4o with the following prompt format:
\vspace{15pt}
\begingroup
\lstset{basicstyle=\scriptsize\ttfamily, frame=single, breaklines=true, breakatwhitespace=true} 

\begin{lstlisting}
In the following, a description of a group of images is provided. 
Summarize the description into a plain, single sentence caption. 
Focus on the most important parts and and keep it as short as possible.
Description: {description}
Caption:
\end{lstlisting}
\endgroup
\vspace{15pt}

Examples on four sets (two from GroupConceptualCaptions and two from GroupWikiArt) are reported in \cref{fig:gc}.

\subsection{Metrics} \label{apx:sec:accuracy_metrics}

For all metrics, we use the suggested standard implementation of recent publications or if available the original authors of the papers suggesting these metrics.
We use the standard \texttt{rouge} package to calculate the ROUGE-L (F1) \cite{lin2004} score.
Furthermore, we use the BLEU-4 score (sentence-wise) \cite{papineni2002} and the METEOR score \cite{denkowski2014} as implemented by the \texttt{nltk} package.
For SPICE \cite{anderson2016} we use the implementation of the \texttt{pycocoevalcap} package.
For CIDEr-D \cite{vedantam2015}, we ported the python 2 code accompanying the original publication \citep{vedantam2015} to python 3.
They also provide precomputed document frequencies on the MSCOCO dataset \citep{lin2014}, which are necessary for our task as only a single reference caption is available.
For BERTScore \cite{zhang2020}, we utilize the \texttt{bert-score} package.
We considered the \texttt{microsoft/deberta-xlarge-mnli} \citep{he2021deberta} model as basis to calculate the scores.
For LLM-as-a-judge \cite{zheng2023}, we utilized the same GPT-4o-mini model as for the rest of the evaluation with the following prompt scheme:
\vspace{15pt}

\lstset{basicstyle=\scriptsize\ttfamily, frame=single, breaklines=true, breakatwhitespace=true}
\begin{lstlisting}
You are an impartial judge evaluating the equivalence of two captions.
Response only with True and False.
Caption 1: {hypothesis}
Caption 2: {reference}
Response:
\end{lstlisting}
\vspace{15pt}

Finally, for CLIPScore \cite{hessel2021} we utilized the Open-CLIP \citep{ilharco2021} ViT-bigG-14 model trained on the LAION2B dataset \citep{schuhmann2022laionb} (S39B B160K).

\subsection{Detailed Results} \label{apx:sec:detailed_results}

We provide detailed results on GroupConceptualCaptions in \cref{tab:apx:groupconcap} and GroupWikiArt in \cref{tab:apx:wikiart}, featuring both all considered metrics as well as the average rank of each method considering all different metrics.
For CLIPScore on GroupWikiArt, it is actually not clear if a good group captioning method should attain higher scores.
Depending on how the vision encoder of CLIP extracts semantic information from the image, it could be rather misleading, \ie, high scores might not indicate a good group caption.
Also, if \eg, trees or bottles are visually important features, generating a description that correctly captures the style, artist and genre might not score high.
Given that we construct the groups from the metadata, we do not foresee that any reference-free metric such as CLIPScore effectively measures performance on this task.
Therefore, we excluded it when calculating the average rank on this dataset, but still report it for reference.

As discussed in the main paper, \texttt{ImageSet2Text} outperforms all baselines on both datasets on average.

\setlength{\tabcolsep}{2pt}
\begin{table*}[ht!]
\caption{Extended results of accuracy evaluation for in group image captioning on the GroupConceptualCaptions dataset. Average rank is calculated over the different metrics. Arrows denote if lower or higher is better.}
\label{tab:apx:groupconcap}
\centering
\begin{tabular}{l|cccccccc|c}
\toprule
\multirow{2}{*}{Model (Setting)}  & CIDEr-D & SPICE & METEOR & Rouge-L & BLEU & BERTScore & LLM-Judge & CLIPScore & Average Rank \\
 & $(\uparrow)$ & $(\uparrow)$ & $(\uparrow)$ & $(\uparrow)$ & $(\uparrow)$ & $(\uparrow)$ & $(\uparrow)$ & $(\uparrow)$ & $(\downarrow)$ \\
\midrule
LLaVA-1.5 (1x1 grid) & 0.138 & 0.079 & 0.101 & 0.163 & 0.019 & 0.644 & 0.233 & 0.270 & 16.00 \\
LLaVA-1.5 (2x2 grid) & 0.145 & 0.115 & 0.101 & 0.150 & 0.020 & 0.657 & 0.276 & 0.271 & 11.31 \\
LLaVA-1.5 (3x3 grid) & 0.075 & 0.132 & 0.085 & 0.115 & 0.014 & 0.657 & 0.336 & 0.287 & 14.56 \\
LLaVA-1.5 (4x4 grid) & 0.058 & 0.122 & 0.088 & 0.109 & 0.014 & 0.659 & 0.336 & 0.281 & 15.06 \\
LLaVA-1.5 (5x5 grid) & 0.066 & 0.115 & 0.083 & 0.107 & 0.013 & 0.652 & 0.284 & 0.279 & 18.44 \\
LLaVA-1.5 (6x6 grid) & 0.066 & 0.121 & 0.084 & 0.099 & 0.013 & 0.651 & 0.302 & 0.274 & 18.06 \\
LLaVA-1.5 (7x7 grid) & 0.077 & 0.111 & 0.083 & 0.102 & 0.014 & 0.653 & 0.293 & 0.269 & 18.00 \\
LLaVA-1.5 (Avg emb.) & 0.078 & 0.052 & 0.091 & 0.130 & 0.019 & 0.608 & 0.121 & 0.234 & 21.25 \\
LLaVA-1.5 (Summary) & 0.099 & 0.107 & 0.115 & 0.115 & 0.016 & 0.632 & 0.138 & 0.301 & 17.25 \\
\midrule
Qwen2.5-VL (1x1 grid) & 0.240 & 0.113 & 0.152 & 0.195 & 0.027 & 0.657 & 0.267 & 0.295 & 7.31 \\
Qwen2.5-VL (2x2 grid) & 0.168 & 0.118 & 0.140 & 0.153 & 0.017 & 0.651 & 0.216 & 0.309 & 11.00 \\
Qwen2.5-VL (3x3 grid) & 0.169 & 0.147 & 0.142 & 0.136 & 0.016 & 0.657 & 0.241 & 0.320 & 7.25 \\
Qwen2.5-VL (4x4 grid) & 0.172 & 0.134 & 0.141 & 0.127 & 0.016 & 0.651 & 0.198 & 0.315 & 10.06 \\
Qwen2.5-VL (5x5 grid) & 0.150 & 0.136 & 0.142 & 0.130 & 0.015 & 0.652 & 0.172 & 0.315 & 10.12 \\
Qwen2.5-VL (6x6 grid) & 0.121 & 0.135 & 0.146 & 0.130 & 0.015 & 0.657 & 0.250 & 0.313 & 10.38 \\
Qwen2.5-VL (7x7 grid) & 0.123 & 0.136 & 0.142 & 0.131 & 0.015 & 0.650 & 0.198 & 0.311 & 11.81 \\
Qwen2.5-VL (Avg emb.) & 0.209 & 0.127 & 0.158 & 0.199 & 0.027 & 0.666 & 0.267 & 0.320 & 4.31 \\
Qwen2.5-VL (Summary) & 0.095 & 0.099 & 0.114 & 0.106 & 0.013 & 0.630 & 0.155 & 0.320 & 19.19 \\
\midrule
GPT-4o (1x1 grid) & 0.251 & 0.130 & 0.137 & 0.189 & 0.024 & 0.655 & 0.302 & 0.299 & 6.81 \\ 
GPT-4o (2x2 grid) & 0.120 & 0.084 & 0.110 & 0.098 & 0.013 & 0.623 & 0.155 & 0.297 & 21.31 \\ 
GPT-4o (3x3 grid) & 0.143 & 0.108 & 0.099 & 0.096 & 0.014 & 0.635 & 0.284 & 0.314 & 15.81 \\ 
GPT-4o (4x4 grid) & 0.146 & 0.105 & 0.104 & 0.099 & 0.013 & 0.649 & 0.276 & 0.315 & 14.81 \\ 
GPT-4o (5x5 grid) & 0.139 & 0.107 & 0.098 & 0.092 & 0.013 & 0.645 & 0.276 & 0.308 & 17.31 \\ 
GPT-4o (6x6 grid) & 0.131 & 0.101 & 0.091 & 0.087 & 0.013 & 0.647 & 0.276 & 0.305 & 19.56 \\ 
GPT-4o (7x7 grid) & 0.112 & 0.106 & 0.097 & 0.091 & 0.013 & 0.649 & 0.259 & 0.294 & 20.00 \\ 
GPT-4o (Summary) & 0.132 & 0.104 & 0.096 & 0.104 & 0.015 & 0.631 & 0.129 & 0.314 & 18.50 \\
\midrule
\texttt{ImageSet2Text} & 0.210 & 0.143 & 0.149 & 0.155 & 0.020 & 0.674 & 0.345 & 0.325 & 2.50 \\ 
\bottomrule
\end{tabular}
\vskip -0.1in
\end{table*}

\setlength{\tabcolsep}{2pt}
\begin{table*}[ht!]
\caption{Extended results of accuracy evaluation for in group image captioning on the GroupWikiArt dataset. CLIPScore is added for completeness, but is not necessarily a suitable metric for this task thus kept gray. It was excluded for computing the average rank over all metrics for GroupWikiArt. Arrows denote if higher or lower is better.}
\label{tab:apx:wikiart}
\centering
\begin{tabular}{l|cccccccc|c}
\toprule
\multirow{2}{*}{Model (Setting)}  & CIDEr-D & SPICE & METEOR & Rouge-L & BLEU & BERTScore & LLM-Judge & \textcolor{gray}{CLIPScore} & Average Rank \\
 & $(\uparrow)$ & $(\uparrow)$ & $(\uparrow)$ & $(\uparrow)$ & $(\uparrow)$ & $(\uparrow)$ & $(\uparrow)$ & \textcolor{gray}{$(\uparrow)$} & $(\downarrow)$ \\
\midrule
BLIP-2 (1x1 grid) & 0.002 & 0.034 & 0.055 & 0.063 & 0.011 & 0.539 & 0.019 & \textcolor{gray}{0.255} & 25.29 \\
BLIP-2 (2x2 grid) & 0.046 & 0.060 & 0.060 & 0.091 & 0.017 & 0.542 & 0.038 & \textcolor{gray}{0.263} & 17.21 \\
BLIP-2 (3x3 grid) & 0.117 & 0.103 & 0.077 & 0.088 & 0.015 & 0.583 & 0.179 & \textcolor{gray}{0.303} & 8.36 \\
BLIP-2 (4x4 grid) & 0.092 & 0.102 & 0.070 & 0.076 & 0.014 & 0.577 & 0.132 & \textcolor{gray}{0.295} & 10.64 \\
BLIP-2 (5x5 grid) & 0.068 & 0.091 & 0.065 & 0.058 & 0.010 & 0.565 & 0.132 & \textcolor{gray}{0.291} & 17.07 \\
BLIP-2 (6x6 grid) & 0.052 & 0.079 & 0.047 & 0.047 & 0.009 & 0.551 & 0.085 & \textcolor{gray}{0.278} & 21.79 \\
BLIP-2 (7x7 grid) & 0.062 & 0.084 & 0.046 & 0.049 & 0.009 & 0.551 & 0.057 & \textcolor{gray}{0.273} & 21.57 \\
BLIP-2 (Avg emb.) & 0.004 & 0.054 & 0.076 & 0.086 & 0.013 & 0.576 & 0.028 & \textcolor{gray}{0.294} & 18.21
 \\
BLIP-2 (Summary) & 0.082 & 0.032 & 0.069 & 0.075 & 0.010 & 0.574 & 0.085 & \textcolor{gray}{0.269} & 17.21 \\
\midrule
LLaVA-1.5 (1x1 grid) & 0.001 & 0.001 & 0.033 & 0.059 & 0.010 & 0.538 & 0.000 & \textcolor{gray}{0.220} & 29.79 \\
LLaVA-1.5 (2x2 grid) & 0.001 & 0.018 & 0.038 & 0.070 & 0.013 & 0.532 & 0.009 & \textcolor{gray}{0.245} & 25.86 \\
LLaVA-1.5 (3x3 grid) & 0.072 & 0.055 & 0.038 & 0.061 & 0.012 & 0.574 & 0.151 & \textcolor{gray}{0.277} & 18.50 \\
LLaVA-1.5 (4x4 grid) & 0.087 & 0.067 & 0.044 & 0.068 & 0.012 & 0.584 & 0.226 & \textcolor{gray}{0.284} & 13.29 \\
LLaVA-1.5 (5x5 grid) & 0.092 & 0.070 & 0.051 & 0.070 & 0.013 & 0.589 & 0.226 & \textcolor{gray}{0.287} & 10.57 \\
LLaVA-1.5 (6x6 grid) & 0.089 & 0.069 & 0.043 & 0.060 & 0.011 & 0.588 & 0.226 & \textcolor{gray}{0.290} & 14.86 \\
LLaVA-1.5 (7x7 grid) & 0.089 & 0.074 & 0.043 & 0.065 & 0.011 & 0.586 & 0.208 & \textcolor{gray}{0.286} & 13.57 \\
LLaVA-1.5 (Avg emb.) & 0.000 & 0.002 & 0.014 & 0.021 & 0.004 & 0.541 & 0.000 & \textcolor{gray}{0.191} & 35.36 \\
LLaVA-1.5 (Summary) & 0.003 & 0.023 & 0.058 & 0.044 & 0.007 & 0.559 & 0.038 & \textcolor{gray}{0.256} & 29.50 \\
\midrule
Qwen2.5-VL (1x1 grid) & 0.001 & 0.031 & 0.074 & 0.061 & 0.007 & 0.576 & 0.057 & \textcolor{gray}{0.252} & 25.00 \\
Qwen2.5-VL (2x2 grid) & 0.004 & 0.053 & 0.089 & 0.060 & 0.007 & 0.583 & 0.123 & \textcolor{gray}{0.284} & 22.29 \\
Qwen2.5-VL (3x3 grid) & 0.007 & 0.064 & 0.087 & 0.057 & 0.007 & 0.595 & 0.179 & \textcolor{gray}{0.295} & 19.64 \\
Qwen2.5-VL (4x4 grid) & 0.015 & 0.070 & 0.097 & 0.065 & 0.007 & 0.597 & 0.189 & \textcolor{gray}{0.294} & 14.71 \\
Qwen2.5-VL (5x5 grid) & 0.014 & 0.068 & 0.103 & 0.065 & 0.008 & 0.599 & 0.274 & \textcolor{gray}{0.304} & 12.86 \\
Qwen2.5-VL (6x6 grid) & 0.013 & 0.073 & 0.101 & 0.065 & 0.008 & 0.595 & 0.236 & \textcolor{gray}{0.303} & 13.00 \\
Qwen2.5-VL (7x7 grid) & 0.008 & 0.069 & 0.097 & 0.065 & 0.008 & 0.595 & 0.217 & \textcolor{gray}{0.300} & 15.29 \\
Qwen2.5-VL (Avg emb.) & 0.000 & 0.042 & 0.099 & 0.062 & 0.008 & 0.593 & 0.075 & \textcolor{gray}{0.285} & 21.21 \\
Qwen2.5-VL (Summary) & 0.000 & 0.033 & 0.064 & 0.027 & 0.004 & 0.566 & 0.057 & \textcolor{gray}{0.272} & 29.57 \\
\midrule
GPT-4o (1x1 grid) & 0.002 & 0.012 & 0.064 & 0.064 & 0.008 & 0.558 & 0.028 & \textcolor{gray}{0.242} & 25.50 \\ 
GPT-4o (2x2 grid) & 0.011 & 0.079 & 0.101 & 0.058 & 0.008 & 0.594 & 0.151 & \textcolor{gray}{0.283} & 16.79 \\ 
GPT-4o (3x3 grid) & 0.023 & 0.117 & 0.116 & 0.050 & 0.007 & 0.613 & 0.208 & \textcolor{gray}{0.300} & 13.43 \\ 
GPT-4o (4x4 grid) & 0.022 & 0.108 & 0.108 & 0.052 & 0.008 & 0.617 & 0.208 & \textcolor{gray}{0.310} & 12.57 \\ 
GPT-4o (5x5 grid) & 0.021 & 0.140 & 0.108 & 0.053 & 0.009 & 0.621 & 0.160 & \textcolor{gray}{0.301} & 12.00 \\ 
GPT-4o (6x6 grid) & 0.032 & 0.177 & 0.108 & 0.050 & 0.008 & 0.624 & 0.142 & \textcolor{gray}{0.305} & 12.57 \\ 
GPT-4o (7x7 grid) & 0.025 & 0.159 & 0.106 & 0.045 & 0.008 & 0.623 & 0.113 & \textcolor{gray}{0.298} & 14.14 \\ 
GPT-4o (Summary) & 0.003 & 0.028 & 0.049 & 0.036 & 0.006 & 0.560 & 0.075 & \textcolor{gray}{0.247} & 29.21 \\
\midrule
\texttt{ImageSet2Text} & 0.032 & 0.063 & 0.115 & 0.090 & 0.012 & 0.620 & 0.248 & \textcolor{gray}{0.291} & 7.57 \\ 
\bottomrule
\end{tabular}
\vskip -0.1in
\end{table*}

\subsection{Experiments using Multi-Image Models}
\label{apx:sec:multi_image}

In addition to the main comparison to baselines as described before, we also compare against VLMs that natively allow to take multiple input images as input for producing a caption.
In particular, we compare the open-source LLaVA-NeXT-Interleave \citep{li2025llavanextinterleave} and the closed source GPT-4o-mini as used throughout this work.

We evaluated those models with a varying number of input images from the same group, which corresponded to the number of images in the grid that we used in the previous experiments.
An important observation is that increasing the number of images incurs a steep increase in compute burden / API costs.
For LLaVA-NeXT-Interleave, putting the full 49 images nearly consumed the whole context window of 32k, thus putting more images into the model would not be possible.
For GPT-4o-mini, we have no way to check how much of the context size is used with 49 images, yet the API requests incur costs of more than 40\$ for the 221 groups we considered in our two benchmarks.
Using the more capable models available through the API would approximately tenfold those costs at the current API pricing (as of October 2025).

The results are shown in \cref{tab:apx:multi_groupconcap} and \cref{tab:apx:multi_wikiart}.
Again, \texttt{ImageSet2Text} outperforms all baselines, achieving a lower average rank across all metrics evaluated.
On GroupWikiArt, LLaVA-NeXT-Interleave (25 images) approaches our performance. However, with slightly fewer or more input images, the gap widens again.
 In fact, we observe no consistent trend that indicates whether increasing or decreasing the number of input images benefits the baselines, suggesting that this hyperparameter is difficult to tune in practice.

\setlength{\tabcolsep}{1.7pt}
\begin{table*}[ht!]
\caption{Results of accuracy evaluation for multi-image VLMs for group image captioning on the GroupConceptualCaptions dataset. Average rank is calculated over the different metrics. Arrows denote if lower or higher is better.}
\label{tab:apx:multi_groupconcap}
\centering
\begin{tabular}{l|cccccccc|c}
\toprule
\multirow{2}{*}{Model (Setting)}  & CIDEr-D & SPICE & METEOR & Rouge-L & BLEU & BERTScore & LLM-Judge & CLIPScore & Average Rank \\
 & $(\uparrow)$ & $(\uparrow)$ & $(\uparrow)$ & $(\uparrow)$ & $(\uparrow)$ & $(\uparrow)$ & $(\uparrow)$ & $(\uparrow)$ & $(\downarrow)$ \\
\midrule
LLaVA-NeXT (4 images) & 0.214 & 0.103 & 0.113 & 0.165 & 0.023 & 0.665 & 0.405 & 0.285 & 5.25 \\
LLaVA-NeXT (9 images) & 0.141 & 0.102 & 0.100 & 0.148 & 0.019 & 0.668 & 0.362 & 0.296 & 7.25 \\
LLaVA-NeXT (16 images) & 0.129 & 0.109 & 0.102 & 0.143 & 0.018 & 0.671 & 0.379 & 0.300 & 6.38 \\
LLaVA-NeXT (25 images) & 0.121 & 0.104 & 0.100 & 0.149 & 0.016 & 0.671 & 0.345 & 0.294 & 7.56 \\
LLaVA-NeXT (36 images) & 0.053 & 0.101 & 0.119 & 0.136 & 0.013 & 0.642 & 0.328 & 0.296 & 10.12 \\
LLaVA-NeXT (49 images) & 0.077 & 0.100 & 0.120 & 0.149 & 0.016 & 0.636 & 0.293 & 0.296 & 8.50 \\
\midrule
GPT-4o (4 images) & 0.155 & 0.111 & 0.120 & 0.135 & 0.018 & 0.650 & 0.250 & 0.309 & 6.88 \\
GPT-4o (9 images) & 0.145 & 0.111 & 0.117 & 0.117 & 0.015 & 0.657 & 0.233 & 0.318 & 8.75 \\
GPT-4o (16 images) & 0.147 & 0.109 & 0.120 & 0.117 & 0.015 & 0.657 & 0.190 & 0.319 & 8.25 \\
GPT-4o (25 images) & 0.159 & 0.116 & 0.120 & 0.121 & 0.015 & 0.656 & 0.207 & 0.320 & 7.19 \\
GPT-4o (36 images) & 0.159 & 0.117 & 0.117 & 0.123 & 0.015 & 0.660 & 0.207 & 0.321 & 6.31 \\
GPT-4o (49 images) & 0.159 & 0.115 & 0.122 & 0.121 & 0.016 & 0.652 & 0.198 & 0.321 & 6.75 \\
\midrule
\texttt{ImageSet2Text} & 0.210 & 0.143 & 0.149 & 0.155 & 0.020 & 0.674 & 0.345 & 0.325 & 1.81 \\
\bottomrule
\end{tabular}
\end{table*}

\setlength{\tabcolsep}{1.7pt}
\begin{table*}[ht!]
\caption{Results of accuracy evaluation for multi-image VLMs for group image captioning on the GroupWikiArt dataset. CLIPScore is added for completeness, but is not necessarily a suitable metric for this task thus kept gray. It was excluded for computing the average rank over all metrics for GroupWikiArt. Arrows denote if higher or lower is better.}
\label{tab:apx:multi_wikiart}
\centering
\begin{tabular}{l|cccccccc|c}
\toprule
\multirow{2}{*}{Model (Setting)}  & CIDEr-D & SPICE & METEOR & Rouge-L & BLEU & BERTScore & LLM-Judge & \textcolor{gray}{CLIPScore} & Average Rank \\
 & $(\uparrow)$ & $(\uparrow)$ & $(\uparrow)$ & $(\uparrow)$ & $(\uparrow)$ & $(\uparrow)$ & $(\uparrow)$ & \textcolor{gray}{$(\uparrow)$} & $(\downarrow)$ \\
\midrule
LLaVA-NeXT (4 images) & 0.004 & 0.012 & 0.039 & 0.067 & 0.013 & 0.552 & 0.009 & \textcolor{gray}{0.230} & 10.14 \\
LLaVA-NeXT (9 images) & 0.022 & 0.054 & 0.064 & 0.065 & 0.013 & 0.587 & 0.104 & \textcolor{gray}{0.268} & 8.14 \\
LLaVA-NeXT (16 images) & 0.063 & 0.069 & 0.078 & 0.099 & 0.017 & 0.606 & 0.104 & \textcolor{gray}{0.283} & 5.14 \\
LLaVA-NeXT (25 images) & 0.069 & 0.069 & 0.087 & 0.097 & 0.017 & 0.609 & 0.113 & \textcolor{gray}{0.292} & 3.86 \\
LLaVA-NeXT (36 images) & 0.026 & 0.067 & 0.049 & 0.059 & 0.007 & 0.561 & 0.123 & \textcolor{gray}{0.288} & 8.00 \\
LLaVA-NeXT (49 images) & 0.022 & 0.058 & 0.050 & 0.057 & 0.007 & 0.548 & 0.104 & \textcolor{gray}{0.275} & 9.43 \\
\midrule
GPT-4o (4 images) & 0.006 & 0.064 & 0.098 & 0.055 & 0.010 & 0.610 & 0.236 & \textcolor{gray}{0.287} & 5.86 \\
GPT-4o (9 images) & 0.010 & 0.065 & 0.085 & 0.039 & 0.007 & 0.606 & 0.208 & \textcolor{gray}{0.285} & 7.86 \\
GPT-4o (16 images) & 0.012 & 0.080 & 0.089 & 0.043 & 0.007 & 0.604 & 0.170 & \textcolor{gray}{0.284} & 6.43 \\
GPT-4o (25 images) & 0.006 & 0.076 & 0.085 & 0.040 & 0.006 & 0.603 & 0.179 & \textcolor{gray}{0.283} & 8.21 \\
GPT-4o (36 images) & 0.019 & 0.082 & 0.087 & 0.038 & 0.006 & 0.608 & 0.179 & \textcolor{gray}{0.290} & 6.50 \\
GPT-4o (49 images) & 0.007 & 0.076 & 0.087 & 0.036 & 0.005 & 0.604 & 0.217 & \textcolor{gray}{0.285} & 8.00 \\
\midrule
\texttt{ImageSet2Text} & 0.032 & 0.063 & 0.115 & 0.090 & 0.012 & 0.620 & 0.248 & \textcolor{gray}{0.291} & 3.43 \\
\bottomrule
\end{tabular}
\end{table*}

\subsection{Compute Infrastructure}
\label{apx:sec:acc_compute}

The experiments investigated in this section were performed on a mix of A40 (48GB) and A100 (80GB) cards, depending on availability on the cluster. Individual nodes featuring 8 GPUs were equipped with 1-2TB of RAM and AMD EPYC CPUs with 64/128 total CPU cores. Calls to OpenAI GPT models were performed using the latest API version as of March 2025. We provide the full code along with all required dependencies. However, the key package versions are as follows:
\begin{itemize}
    \item \texttt{python = 3.10}
    \item \texttt{pytorch = 2.6.0+cu124}
    \item \texttt{transformers = 4.49}
\end{itemize}

\section{Completeness Evaluation}
\label{sec:completeness_supp}

To assess the completeness of our descriptions, we conduct an experiment on the downstream task of Set Difference Captioning using the PairedImageSets dataset \cite{visdiff}.

\subsection{Integrating \texttt{ImageSet2Text} into VisDiff}
\label{apx:graph2visdiff}
As outlined in the main paper, the proposer-ranker framework introduced in VisDiff begins with single-image captions generated via BLIP-2 on a randomly selected subset of each of the two considered sets. In our experiment, we instead consider the information extracted through \texttt{ImageSet2Text} as a starting point of the same proposer-ranker framework. Below, we detail the key implementation aspects of this experiment.

First, the input provided to the proposer to identify differences between the sets is the concept graph representations generated at the final iteration of \texttt{ImageSet2Text} for both image sets $\mathcal{D}_A$ and $\mathcal{D}_B$. 
The concept graphs are transformed into a textual format using the \texttt{generate\_network\_text} function from the NetworkX library in Python\footnote{NetworkX, \url{https://networkx.org/documentation/stable/reference/readwrite/generated/networkx.readwrite.text.generate_network_text.html}, Last Access: 7th of March 2025.}.

In the original VisDiff implementation, the authors conduct three rounds of their pipeline, where, in each round, 10 different images are considered to generate 10 candidate differences. The proposed differences are merged over the three rounds (for a total of 30 differences) and then passed as input to the ranker module. Since our concept graphs are precomputed and remain static, we have considered using a single round of iteration. However, we observed that when the proposer is prompted to find 30 possible differences at once, the proposals start becoming meaningless, diverging toward irrelevant interpretations. To mitigate this, we adopt a two-round approach, requesting 15 differences per round, for a total of 30 proposed differences.

While the proposer's prompt remains largely unchanged, we make one key modification: instead of specifying that the input consists of 10 individual captions, we explicitly clarify that the input consists of descriptions of two image sets, represented in graph form.

\subsection{Failure cases of VisDiff}
\label{apx:com:fails}
\begin{figure*}
    \centering
    \includegraphics[width=\linewidth]{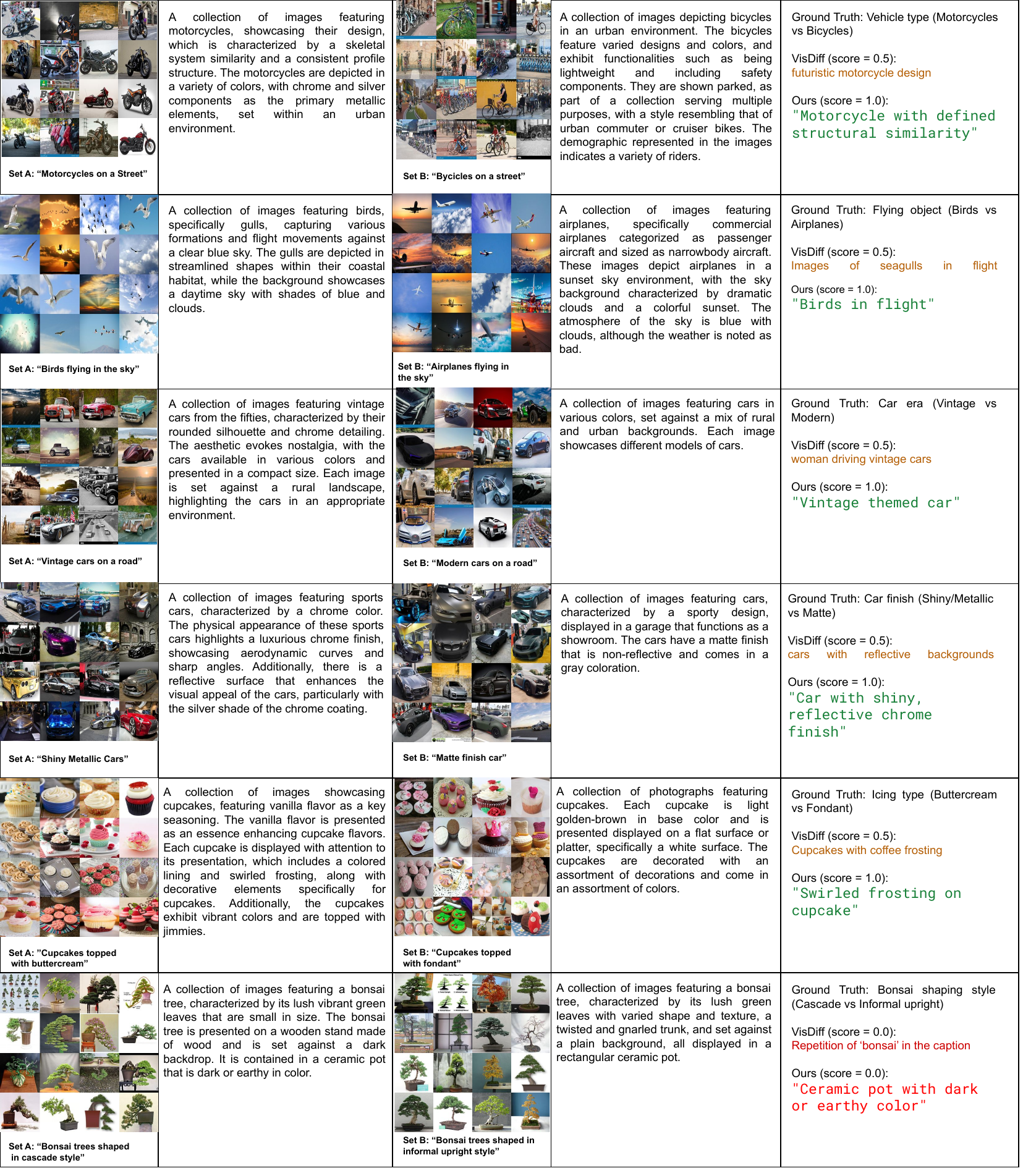}
    \caption{Case-by-case comparison with VisDiff. The first column presents images of ``Set A" along with their definition in PIS, the second column presents our description for this set. The third column presents images of ``Set B" and their definition in PIS, while the fourth column is our generated description. Finally, in the fifth column we report the ground truth of the difference between Set A and Set B, the prediction and score of VisDiff and our prediction and score.}
    \label{fig:visdiff_comparison}
\end{figure*}

In the main paper, we demonstrated that integrating the information extracted through \texttt{ImageSet2Text} enhances performance on the PairedImageSets dataset. In this section, we further examine this improvement through a case-by-case analysis, directly comparing our results with those of VisDiff on six specific failure cases reported in their paper \cite{visdiff}. The comparison is illustrated in \cref{fig:visdiff_comparison}.

As noted by the authors of VisDiff, one of the primary limitations of their approach is that the BLIP-2-generated captions tend to be overly generic. This issue is particularly evident in more challenging cases, where a deeper, more nuanced understanding of the images is required, such as distinguishing between “Cupcakes topped with buttercream” and “Cupcakes topped with fondant”. In contrast, \texttt{ImageSet2Text} addresses this limitation by iteratively refining the focus of the VQA, ensuring that the generated descriptions capture more specific and contextually relevant details. As shown in \cref{fig:visdiff_comparison}, our experimental setting produces superior set difference captions in five out of the six cases.

\subsection{Compute Infrastructure}
\label{apx:com:compute}

The experiments investigated in this section were conducted in Google Colab using a Tesla T4 GPU (16GB VRAM). Calls to OpenAI GPT models were performed using the latest API version as of March 2025. We provide the full code of our experiments, with the most important packages being:
\begin{itemize}
    \item \texttt{python = 3.11}
    \item \texttt{pytorch = 2.6.0+cu124}
    \item \texttt{nltk = 3.9.1}
\end{itemize}

\section{User Study}
\label{sec:user_supp}

\begin{figure}
    \centering
    \includegraphics[width=\linewidth]{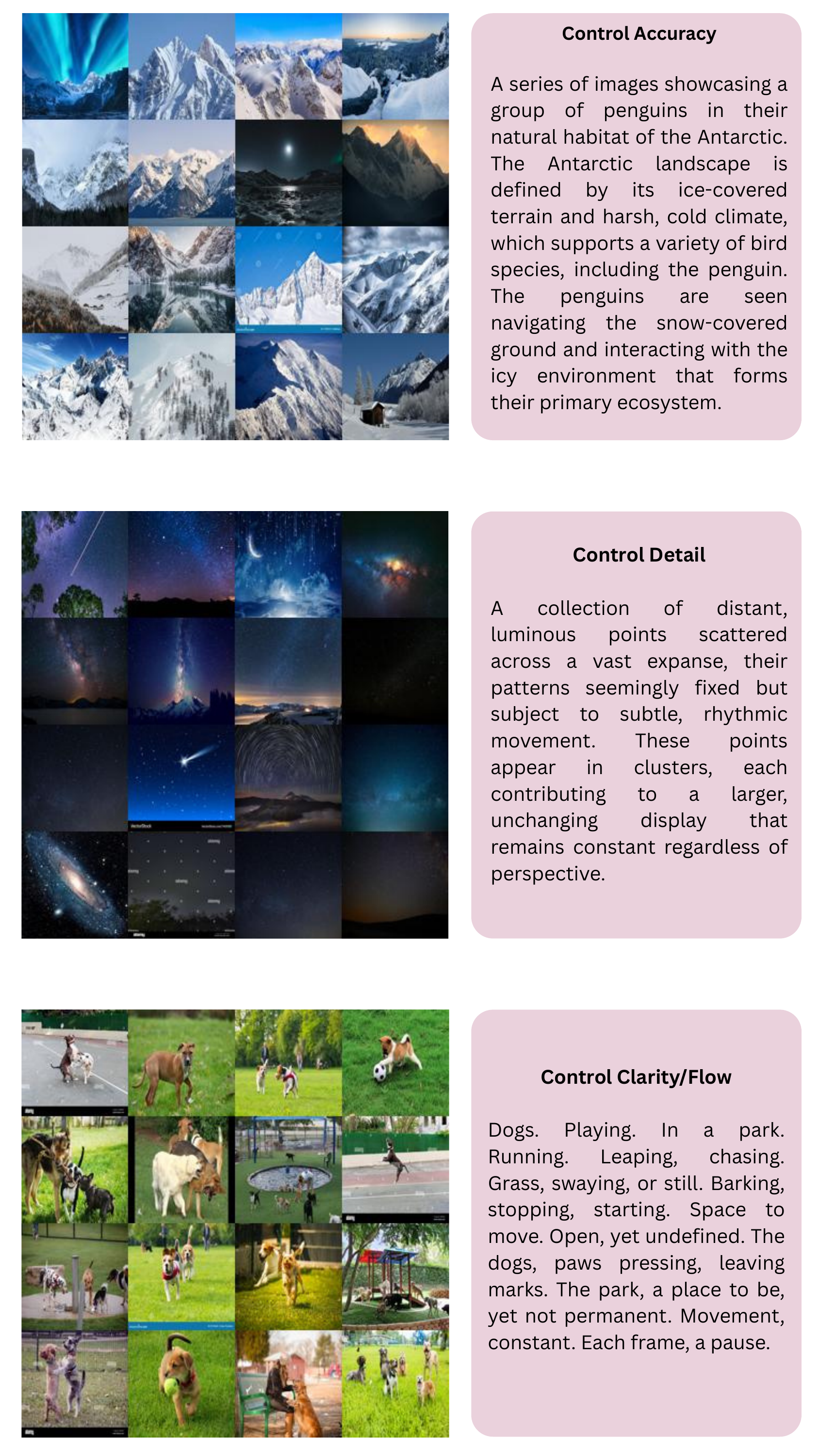}
    \caption{Three examples of control descriptions used as reference values for the user study.}
    \label{fig:control}
\end{figure}

\begin{figure}
    \centering
    \includegraphics[width=\linewidth]{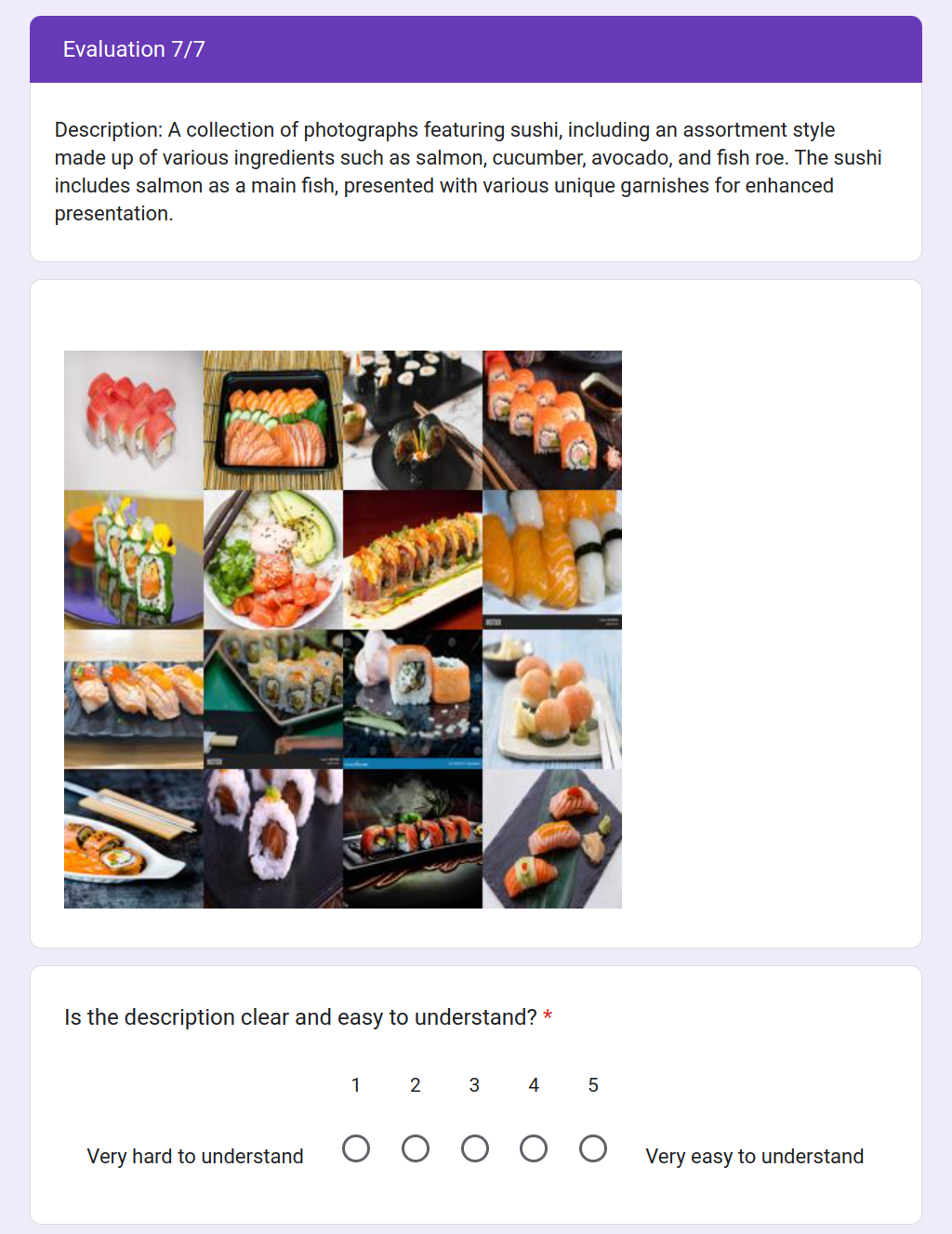}
    \caption{Example of a pair set-description shown in the user study, along with the first rating question about clarity.}
    \label{fig:interface}
\end{figure}

To assess the quality of descriptions generated by \texttt{ImageSet2Text}, we conducted a user study with 233 participants recruited via Prolific. The study was implemented as a dynamic Google Form, deployed as a Google Web Application, and coded using Google Apps Script. Each user evaluates seven descriptions: six generated by \texttt{ImageSet2Text} and one control description. Three examples of control descriptions (one per type) and their corresponding image grids are shown in \cref{fig:control}. The study takes approximately 8 minutes per user. The user study questions are listed in \cref{tab:user_study} and a screenshot of the interface is reported in \cref{fig:interface}, where a pair set-description is reported, along with the first question of the user study. 

\begin{table}
\caption{Questions of the user study. Each question allows answers on a Likert scale from 1 to 5.}
\label{tab:user_study}
\centering
\begin{tabular}{p{\linewidth} p{0.65\linewidth}}
\toprule
Question \\
\midrule
(1) Is the description clear and easy to understand? \\
(2) Does the description contain enough details? \\
(3) Does the description contain misleading or incorrect information? \\
(4) Does the text flow naturally? \\
(5) What is your overall satisfaction for this description? \\
\bottomrule
\end{tabular}
\vskip -0.1in
\end{table}

\section{Failure cases}
\label{apx:failures}

In this section, we detail the failure cases caused by the integration of WordNet and CLIP in the verification process of \texttt{ImageSet2Text}, along with an experimental analysis to estimate their frequency and impact on the quality of the generated descriptions. We also discuss additional limitations inherited from the individual components that constitute the method’s pipeline.

\begin{figure*}[ht]
    \centering
    \begin{subfigure}[b]{\textwidth}
        \centering
        \includegraphics[width=\linewidth]{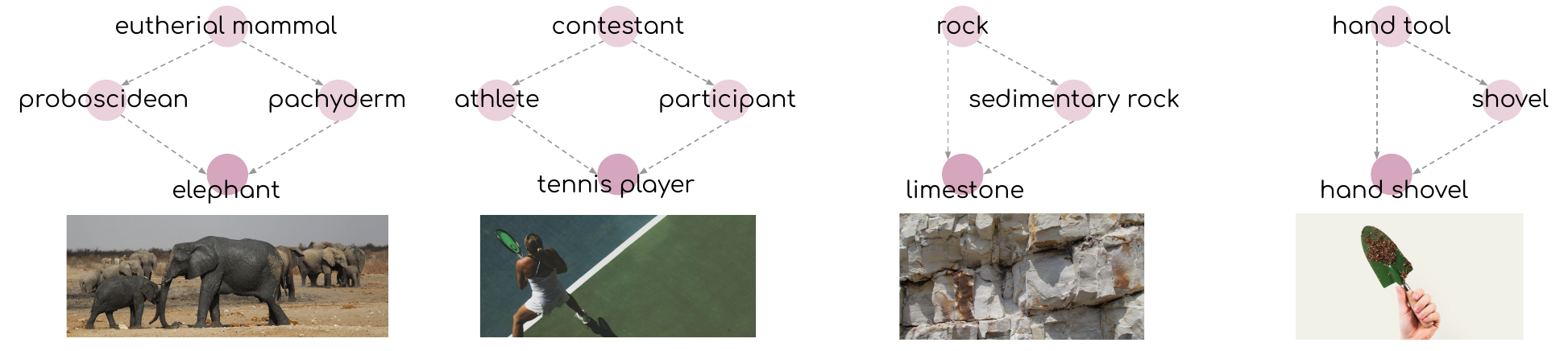}
        \caption{Four examples of nodes with siblings that are not mutually exclusive. From left to right, the first two fall into the \textbf{rhomboid case} and the second two fall into the \textbf{triangular case}.}
        \label{fig:mutual_exclusiveness_a}
    \end{subfigure}
    
    \vspace{0.5cm}

    \begin{subfigure}[b]{\textwidth}
        \centering
        \includegraphics[width=\linewidth]{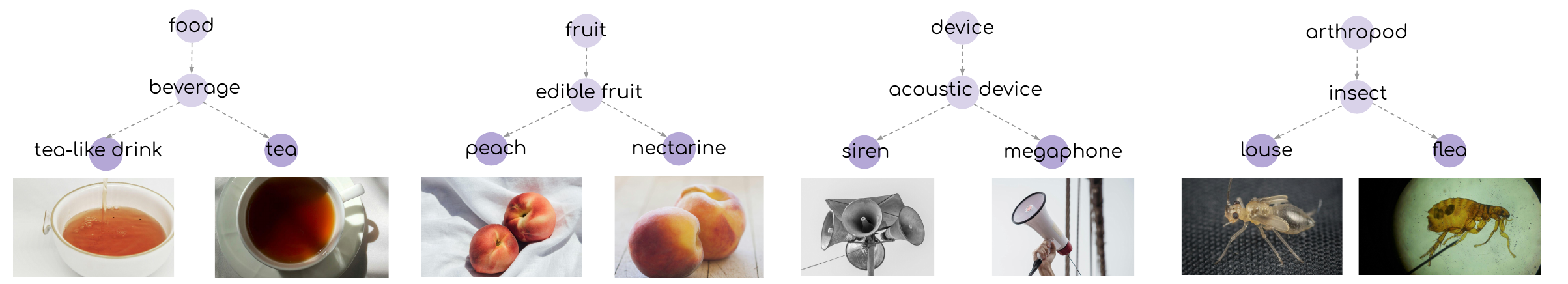}
        \caption{Four examples of sibling nodes that are semantically distinct but visually too similar to be effectively distinguished in the CLIP embedding space.}
        \label{fig:mutual_exclusiveness_b}
    \end{subfigure}
    
    \caption{Illustration of failure cases involving non-mutually-exclusive or visually similar sibling nodes in WordNet.}
    \label{fig:mutual_exclusiveness_combined}
\end{figure*}

\subsection{Non-mutually exclusive sibling nodes}

As described in Section 3 of the main paper, given a hypothesis $h_i \in H$ to verify, the set of contradicting hypotheses $\mathcal{H}_i^-$ is constructed by replacing the object $o_i$ in $h_i$ with its sibling nodes. This approach assumes that sibling nodes are mutually exclusive: a condition that does not always hold in WordNet. Some representative counterexamples are shown in Fig. \ref{fig:mutual_exclusiveness_a}, which illustrates two distinct structural patterns:
\begin{itemize}
    \item \textbf{Rhomboid case:} A node has two hyponyms that share a common descendant. For example, \textit{proboscidean} and \textit{pachyderm} are sibling nodes, but both have \textit{elephant} as a hyponym, and are therefore not mutually exclusive. A similar structure is seen with \textit{tennis player}, which is a child of both \textit{participant} and \textit{athlete}.
    \item \textbf{Triangular case:} A lexical concept can be linked to a hypernym at different levels of granularity. For instance, \textit{limestone} is a child of both \textit{rock} and \textit{sedimentary rock}, where the latter is itself a child of \textit{rock}. Consequently, while \textit{limestone} and \textit{sedimentary rock} appear as siblings under \textit{rock}, they are also hierarchically connected and not mutually exclusive. A similar case is observed with \textit{hand shovel}, as shown in the figure.
\end{itemize}

The existence of these structural patterns poses challenges for the verification phase. Therefore, it is important to estimate how many nodes fall under this condition to its impact in the definition of the set $\mathcal{H}_i^-$. To estimate the number of affected nodes, we define as \textbf{intersection leaf nodes} those that have multiple hypernyms in the WordNet hierarchy. Given an intersection node, we trace upwards through the hierarchy to identify their lowest common ancestor, which we refer to as the \textbf{intersection root node} (\eg, \textit{eutherial mammal} is the intersection root node for the intersection leaf node \textit{elephant}).

The verification issue arises for the hyponyms of an intersection root node, since at least two of them lead to the intersection leaf node, hence representing concepts that are not mutually exclusive. We compute the total number of intersection root nodes in WordNet and compare that to the total number of nodes eligible for verification (\ie, all non-leaf nodes). Our analysis shows that only 1.49\% of the nodes are affected by this issue. This low percentage supports the use of the current definition of $\mathcal{H}_i^-$ in our extensive experiments, where it has not posed significant problems. However, despite its rarity, this failure case should be acknowledged in future iterations of this work, where more robust strategies could be considered.

\subsection{Visual similarity among siblings}

The second failure case that we identified also affects the verification phase. Specifically, it involves the kNN classifier operating over sibling nodes that represent mutually exclusive concepts which are not visually distinguishable. Representative examples are shown in Fig.~\ref{fig:mutual_exclusiveness_b}. In such cases, the kNN classifier may struggle to differentiate between the hypothesis $h_i$ under verification and its contradicting hypotheses in $\mathcal{H}_i^-$. While this limitation is acceptable in general (\emph{i.e.}, the system only needs to be as sensitive as the visual data allows), it becomes an issue when these indistinguishable siblings are embedded within a broader class that includes clearly distinct concepts.

For instance, suppose we are verifying the hypothesis that a set of images depicts a \textit{tea-like drink}, but the hypothesis is rejected due to the presence of \textit{tea} among the contradicting alternatives, which is visually inseparable from \textit{tea-like drink} in the CLIP embedding space. As a result, the verification process halts at a higher-level concept in the lexical hierarchy, such as \textit{beverage}, which is too generic to capture the intended meaning. This loss of granularity leads to the generation of overly vague descriptions, even when the visual evidence could support more precise interpretations.

To assess the severity of this issue, we perform an empirical estimate of how frequently this failure case occurs and how significantly it impacts the verification process. Unlike the previous case, this analysis requires access to visual information, which is not provided by WordNet. Interestingly, ImageNet~\cite{deng2009} builds its classes based on the WordNet nouns hierarchy, making it a suitable resource for investigating visually indistinguishable sibling nodes. It is important to note that not all WordNet nodes are present in ImageNet, as some concepts are inherently non-visual and therefore excluded. For our analysis, we rely on a subset of ImageNet, assuming that the resulting percentage of affected cases is representative of what would be observed over the full dataset. In particular, over the total of 16693 nodes eligible for verification (\ie the non-leaf nodes), only 5991 has at least one hyponym present in ImageNet. Because we had access to a reduced version of ImageNet, we consider 874 of these nodes.

For each node (\eg \textit{beverage}) we consider all possible hyponym pairs, which can be either \textbf{visually distinguishable} (\eg \textit{tea} and \textit{coffee}) or \textbf{visually indistinguishable} (\eg \textit{tea} and \textit{tea-like drink}). We define a pair visually distinguishable if, given five images per hyponym, CLIP correctly matches at least 8 out of 10 images to their respective hyponym, based on similarity scores. Conversely, we define the pair visually indistinguishable if this condition does not hold. In addition, we consider a node as \textbf{overgeneric} if the proportion of visually indistinguishable pairs of its hyponyms falls below a predefined threshold. Formally, a node is marked as overgeneric if:
\[
\frac{\#_{\text{indist}}}{\#_{\text{dist}} + \#_{\text{indist}}} < \phi
\]
where:
\begin{itemize}
  \item $\#_{\text{dist}}$ is the number of visually distinguishable hyponym pairs of the node,
  \item $\#_{\text{indist}}$ is the number of visually indistinguishable hyponym pairs of the node,
  \item and $\phi$ is a predefined threshold.
\end{itemize}

Using this criterion, and setting $\phi = 0.15 $ we identified 51 unreliable nodes, corresponding (in proportion) to $2.09\%$ of the total non-leaf nodes of WordNet, providing an estimate of the prevalence of this specific failure case. By setting the threshold $\phi$ to a low value, we capture situations where failures emerge not from CLIP's inherent weaknesses (discussed in Section E.3), but from the presence of concepts actually indistinguishable visually.

Given the low occurrence of this failure case, its impact is negligible in the current version of \texttt{ImageSet2Text}. However, in future iterations of this work, it may be beneficial to account for such ambiguity. One possible solution could involve introducing \textbf{disjunction nodes}, \ie allowing \texttt{ImageSet2Text} to express uncertainty through formulations like \textit{tea or tea-like drink}.

\subsection{Other Limitations}

In addition to the previously discussed failure cases, primarily arising from the integration of WordNet and CLIP during the verification phase, we identify further limitations of \texttt{ImageSet2Text} that stem from the individual components of the pipeline.

One notable constraint concerns the language model: \texttt{ImageSet2Text} currently depends on GPT-4o-mini, a proprietary model. In future work, we aim to investigate open-source alternatives to enhance the accessibility of the pipeline for researchers and users with limited resources.

As a pipeline that heavily relies on large language models, \texttt{ImageSet2Text} also inherits many of their well-documented limitations. These include a tendency to generate plausible but incorrect content, sensitivity to prompt phrasing, and difficulties with factual consistency and reasoning in complex tasks \citep{bommasani2021opportunities}. Broader challenges, such as opacity, and the risk of biased or unreliable outputs, are especially salient in foundation models \citep{liang2023holistic}. Our prompt engineering efforts were devoted to mitigate these limitations as much as possible. As research on LLMs progresses and these limitations are addressed by the community, also the quality of the descriptions generated by \texttt{ImageSet2Text} is expected to improve.

We also highlight limitations related to the usage of CLIP as a zero-shot classifier for verification. While CLIP has proven remarkably effective for this task over broad and generic categories, its performance degrades significantly in fine-grained and open-world settings \cite{bianchi2024clip}. A central limitation lies in CLIP's difficulty distinguishing between semantically or visually similar classes, especially when these distinctions are subtle or domain-specific. This stems from both the model’s reliance on internet-scale, weakly labeled data and its coarse-grained alignment between images and text. These limitations suggest that CLIP, while powerful, remains a bottleneck for open-world applications of \texttt{ImageSet2Text}.

Another limitation we encountered arises during the hypothesis expansion phase, where reliance on WordNet introduces ambiguity due to polysemy: \ie, the presence of multiple definitions for the same word, as is common in any lexical resource. In some cases, this can lead to mismatches between the intended definition of a word and the WordNet selected hypernyms for generalizing the hypothesis $o_0$ in $h_0$ and generating the set $\mathcal{H}$. We were able to mitigate this issue effectively by prompting the LLM to select the most contextually appropriate WordNet node, based on $o_0$. This simple yet effective strategy allowed us to maintain semantic precision in the expansion process without requiring manual intervention.

\section{Automatic Alternative Text Generation of Image Sets}
\label{sec:app_supp}

Automatic generation of alternative text (alt-text) is a fundamental application of image captioning \cite{gurari2020}, particularly for visually impaired individuals. \texttt{ImageSet2Text} introduces a novel approach by summarizing entire image collections rather than generating captions for individual images. Since this area is largely unexplored, we conducted interviews with three collaborators from the Spanish national association for universal accessibility,  Fundación ONCE, to gather community feedback on its usefulness and potential improvements \cite{costanza2020}. The interviews were conducted by the first author in Spanish. They were transcribed and then carefully translated into English to facilitate a qualitative analysis of the collected data. The questions are summarized in \cref{tab:focus_questions}. 

\begin{table*}
\caption{Focus Areas and Associated Questions to explore the usability of \texttt{ImageSet2Text} in the context of alternative text generation.}
\label{tab:focus_questions}
\centering
\begin{tabular}{p{0.3\linewidth} p{0.65\linewidth}}
\toprule
Focus & Questions \\
\midrule
Alt-Text Generation Usage & Do you use any tool or service for generating image descriptions? If so, which ones? \\
 & In what contexts do you find image descriptions most useful? (\textit{e.g.}, on websites, social media, documents) \\
\midrule
\texttt{ImageSet2Text} opportunities & If you had the option to receive a summary of a set of images instead of individual descriptions, do you think it would be useful? Why or why not? \\
 & In what situations or types of content do you think this option would be most beneficial? (\textit{e.g.}, news articles, academic documents, presentations, social media, personal images from events/travel, etc.) \\
 & Can you think of specific cases where a summary of a set of images would be more useful than individual descriptions? \\
 & Do you think these types of summaries could be useful in professional, educational, or personal contexts? How? \\
 & Do you see any difficulties with this approach? \\
\midrule
\texttt{ImageSet2Text} evaluation & We have developed a method to automatically create descriptions of image sets: would you like to review an example and share your feedback? \\
& What aspects of the descriptions do you find clear or useful? \\
 & Are there any parts of the descriptions that you find confusing or unclear? \\
& How could we improve the structure, level of detail, or language used in the descriptions to make them more accessible? \\
& Would you prefer these summaries to be presented in a specific format? (\textit{e.g.}, structured lists, narrative summaries, bullet points) \\
\midrule
Extra & Is there anything we haven't mentioned that you think we should consider when designing this methodology? \\
\bottomrule
\end{tabular}
\vskip -0.1in
\end{table*}

In the first part of the interview, the questions aimed to assess whether the interviewee was familiar with tools for automatic alt-text generation. The second part explored the potential usefulness of accessing textual descriptions for collections of images in various tasks. In the third part, we presented an example of description generated by \texttt{ImageSet2Text} and asked the interviewee to provide feedback on it. Finally, the interview concluded with an open-ended opportunity for the interviewee to share any additional relevant information.

Our collaborators generally welcomed the idea, recognizing that set-level descriptions could be extremely useful when understanding the broader context of a scene is more important than focusing on specific details, such as during events and entertainment, keeping memories of travels, or while managing folders on the computer. However, they emphasized that such summaries should complement rather than replace individual image descriptions, as both serve distinct purposes. When evaluating an example description, participants expressed overall satisfaction with the level of detail, coherence, and clarity. In particular, they appreciated explicit relations among the entities in the images, an aspect they often find lacking in commercial automatic alt-text generators. This feature of \texttt{ImageSet2Text} is a direct consequence of integrating structural representations that explicitly consider relationships between visual elements \cite{phueaksri2023,phueaksri2024}.

Should \texttt{ImageSet2Text} be further developed in the context of accessible technologies, our collaborators from Fundación ONCE suggested key areas for improvement. First, they emphasized the importance of using simple, clear, and direct language to minimize ambiguity. They also recommended tailoring descriptions based on the user’s visual experience, \eg those who have seen before might benefit from references to colors and light, while those who have never seen may require alternative descriptions. Another point raised was that the current approach works best for homogeneous image sets with shared visual elements. In real-world scenarios, however, image collections might be more heterogeneous. As a result, a necessary future direction for \texttt{ImageSet2Text} is to not only identify common features but also detect distinct clusters within an image set to provide more meaningful summaries, aligning with ongoing research in semantic image clustering \cite{liu2024}.

Overall, this feedback highlights the potential of \texttt{ImageSet2Text} to enhance accessibility and inclusion for blind and low-vision users in both personal and professional settings.   

\end{document}